\RequirePackage{fix-cm}
\documentclass[smallextended]{svjour3}       %
\usepackage[utf8]{inputenc}

\smartqed  %
\usepackage{graphicx}
\usepackage[hidelinks,colorlinks=false,urlcolor=blue,linkcolor=black]{hyperref}
\usepackage{mathtools}
\usepackage{subcaption}
\usepackage{algpseudocode}
\usepackage{bm}
\usepackage{amsfonts}
\usepackage{amsmath}
\usepackage{amssymb}
\usepackage{pdfpages}
\usepackage{multicol}
\usepackage{natbib}

\DeclareMathOperator*{\argmax}{arg\,max}
\DeclareMathOperator*{\argmin}{arg\,min}

\journalname{Data Mining and Knowledge Discovery}
\begin{document}

 \thispagestyle{empty}
\parbox{\textwidth}{
This is an author manuscript of the following publication: \smallskip
\\ Georg Krempl, Dominik Lang, Vera Hofer. 
\\ Temporal Density Extrapolation using a Dynamic Basis Approach.
\\ In: K. Borgwardt, Po-Ling Loh, Evimaria Terzi, Antti Ukkonen (eds.). 
\\ Data Mining and Knowledge Discovery 
\\ Special Issue of the ECML/PKDD 2019 Journal Track, Springer, 2019. 
\\ The original publication is available at \url{https://link.springer.com/journal/10618/} 
}
\newpage \setcounter{page}{1}

\title{Temporal Density Extrapolation using a Dynamic Basis Approach}
\titlerunning{Temporal Density Extrapolation}

\author{G. Krempl$^1$        \and
        D. Lang \and
        V. Hofer %
}
\authorrunning{G. Krempl, D. Lang, V. Hofer}

\institute{G. Krempl \at
              Utrecht University, The Netherlands.
              \email{g.m.krempl@uu.nl}           %
           \and
           D. Lang \at
              Otto-von-Guericke University Magdeburg, Germany.
              \email{dominik.lang@ovgu.de}           %
           \and
           V. Hofer \at
              Karl-Franzens-University Graz, Austria.
              \email{vera.hofer@uni-graz.at}           %
}

\date{\today}

\date{Received: \today / Accepted: date}

\maketitle
\footnotetext[1]{All authors contributed equally to this publication.}

\begin{abstract}
Density estimation is a versatile technique underlying many data mining tasks and techniques, ranging from exploration and presentation of static data, to probabilistic classification, or identifying changes or irregularities in streaming data. With the pervasiveness of embedded systems and digitisation, this latter type of streaming and evolving data becomes more important. Nevertheless, research in density estimation has so far focused on stationary data, leaving the task of of extrapolating and predicting density at time points outside a training window an open problem.
For this task, Temporal Density Extrapolation (TDX) is proposed. This novel method models and predicts gradual monotonous changes in a distribution. It is based on the expansion of basis functions, whose weights are modelled as functions of compositional data over time by using an isometric log-ratio transformation. Extrapolated density estimates are then obtained by extrapolating the weights to the requested time point, and querying the density from the basis functions with back-transformed weights.
Our approach aims for broad applicability by neither being restricted to a specific parametric distribution, nor relying on cluster structure in the data. It requires only two additional extrapolation-specific parameters, for which reasonable defaults exist. Experimental evaluation on various data streams, synthetic as well as from the real-world domains of credit scoring and environmental health, shows that the model manages to capture monotonous drift patterns accurately and better than existing methods. Thereby, it requires not more than $1.5$ times the run time of a corresponding static density estimation approach.


\end{abstract}

\section{Introduction}

The extent and scope of available data continues to grow, often comprising data that is continuously generated over longer time spans, in environments that are subject to changes over time \citep{ReinselGantzRydning2017}. 
Volume and velocity of such data streams often require automated processing and analysis, while their dynamic nature and associated volatility needs to be considered as well \citep{FanBifet2013}.
For example, the distribution of the data might change over time, a problem that is commonly denoted as concept drift \citep{WidmerKubat1996} or population drift \citep{KellyHandAdams1999}. In prediction tasks, such drift requires adaptation mechanisms, for example by forgetting outdated irrelevant data or models \citep{GamaEtal2014}. 
However, merely attempting to avoid the negative influence that this dynamic nature can have on statistical models is leaving its potential unused. In contrast, aiming to understand and to incorporate the changes in data into the statistical model itself might provide additional value, by helping to perform more accurate predictions for the future and allowing a structured description of the occurring changes.

Several tasks and approaches have been proposed to identify change or irregularities in data, such as outlier detection (surveyed for example in \citep{SadikGruenwald2014}), anomaly detection (surveyed by \citep{ChandolaBanerjeeKumar2009}), or change detection (surveyed by \citep{TranGaberSattler2014}). Furthermore, some research has focused on the exploration, understanding, and exploitation of change: change diagnosis aims to estimate the spatio-temporal change rates in kernel density within a time window  \citep{Aggarwal2005}. This might be seen as an early proponent of the change mining paradigm, which was proposed afterwards by \citep{BottcherHoppnerSpiliopoulou2008} and calls for data mining approaches that provide understanding of change itself. 
Following this paradigm, drift mining techniques aim to provide explicit models of distributional changes over time, for example to transfer knowledge between different time points in data streams under verification latency \citep{HoferKrempl2012A}.

An essential technique in data science \citep{ChaconDuong2018,Scott2015} that is underneath many of the algorithms in the tasks above is density estimation \citep{Rosenblatt1956,Whittle1958}, which is also important for exploration and presentation of data in general \citep{Silverman1986}, as it is for probabilistic classifiers such as Parzen Window Classifiers \citep{Parzen1962}.
Given a set of instances sampled within a training time window from a non-stationary (i.e., drifting) data stream, the provided density estimates typically correspond to the distribution over all instances within this window. However, these estimates might also be required for specific time points, rather than time windows. This time point might lie within the training time window, requiring an in-sample interpolation.
Alternatively, it might lie outside, requiring an out-of-sample extrapolation to past or future time points.

Research in density estimation has so far focused on providing in-sample estimates, which are often adapted to the distribution of the most recently observed samples in the training time window. In contrast, density extrapolation beyond this window has only recently received attention \citep{Krempl2015}, although it has some potential beyond being simply used in lieu of static density estimations in the applications above:
Most importantly, extrapolating continuing trends allows to anticipate future distributional changes and to take preparatory measures. Examples are classifiers or models that incorporate forthcoming distributional changes anticipatively; 
active learning and change detection approaches that proactively sample in regions where change is expected to happen; or in the identification of {unexpected} changes in drift.

We propose a novel approach to model and predict the density of a feature in a data stream. The underlying distribution is described as an expansion of Gaussian density functions, which are placed at fixed equidistant locations\footnote{This is related to Gaussian Mixture Models (GMM), with the subtle but important difference that in GMMs the location and variance parameters of the components are in general estimated from data and not fixed a priori. Note that the model is not restricted to Gaussian basis functions.}. The basis weights of the Gaussian density functions are fitted functions of time. In this way the changes observed in the data over time are modelled as a change in the weighting of the basis expansion.\\

\begin{figure}[h]
\centering
\includegraphics[width=\textwidth]{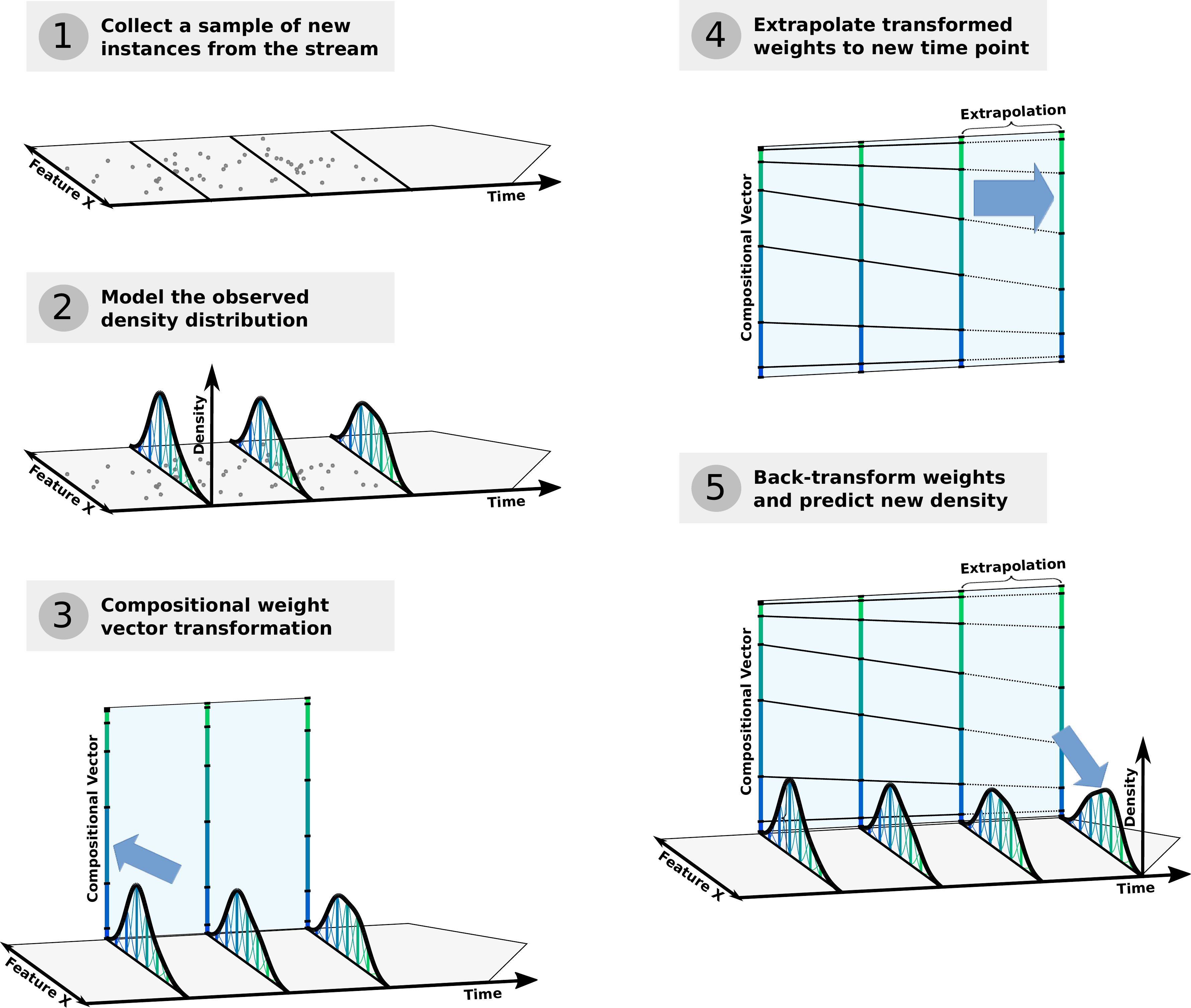} 
\caption{Illustration of Temporal Density Extrapolation, which aims to capture and predict the change in the distribution of instances over time. 
(1) A sample of instances $X$ is observed at time points $T$ in a data stream. 
(2) The observed density distribution is modelled as an expansion of density functions.
(3) Their basis weights form a compositional vector. 
(4) The weight development over time is modelled, and extrapolated to new time points.
(5) These extrapolated weights are back-transformed to predict the resulting new density distribution.
}\label{fig:graphical_abstract}

\end{figure}

Fig. \ref{fig:graphical_abstract} shows a feature $X$ in a data stream, whose density at different points in time is described by an expansion of six Gaussian density functions. The weights associated to each of these basis functions form a composition \citep{Aitchison1982, EgozcueEtal2003}   
that sums to one, their proportions are illustrated on the back end of the figure. The change in the density at the second and the third time point is modelled as a change in the distribution of the weights, as visible from the shift in the composition proportions. It is the core of the approach's fitting process to model these weights as functions of time, which will be elaborated on in Sec. \ref{sec:method}. 

Modelling the data stream in this way entails less computation compared to kernel density estimators, since we do not need to complete a full kernel matrix but only evaluate the small number of basis functions at the sample positions. This approach also allows a straight-forward way of forecasting the density at future time points, since the basis weight functions only need to be extrapolated to the desired time. Furthermore, the model delivers an easily interpretable description of drift by means of the basis weight functions.

The remainder of this article is organised as follows:
In the section \ref{sec:relwork}, we review the related work, before presenting and discussing our temporal density extrapolation approach in detail in section \ref{sec:method}. An experimental evaluation of this approach is given in section \ref{sec:expeval}, concluding remarks in section \ref{sec:conc}.

\section{Related Work} \label{sec:relwork}
There is a vast literature on estimating the density within a training sample \citep{ChaconDuong2018,Scott2015}, 
This includes approaches that provide spatio-temporal density estimates for time points within the training time window, e.g., by using time-varying mixture models \citep{LawlorRabbat2016} or spatio-temporal kernels \citep{Aggarwal2005}.
However, our focus is on the task of density extrapolation in time-evolving data streams, as recently formulated in \citep{Krempl2015}. Thus, we will restrict the following review to this density extrapolation, then review the related problem of out-of-sample density forecasting, and finally discuss the broader context within the literature of concept drift in data streams.

Density extrapolation is described in \citep{Krempl2015}, together with a sketch of the general idea to extend kernel density estimation techniques for this task. In \citep{Lampert2015}, ``Extrapolating Distribution Dynamics'' (EDD) is proposed, although this approach aims for predicting the distribution in the \emph{immediate} future one-step ahead. EDD models the transformation between previous time points and applies this transformation to the most recent sample, to obtain a one-step ahead extrapolation of distribution dynamics.

A related problem is density forecasting (see, e.g., the survey in \citep{TayWallis2000}), where the realisations of a random variable are predicted. 
Applications exist in macroeconomics and finance \citep{TayWallis2000,Tay2015}, as well as in specialised fields like energy demand forecasting \citep{HeLi2018,MokilaneEtal2018}. 
Forecast are either within (in-sample) or outside (out-of-sample) an observed sample and training time window. In our context, only out-of-sample forecasting is relevant. 
In \citep{GuHe2014}, the 'dynamic kernel density estimation' in-sample forecasting approach by \citep{HarveyOryshchenko2012} is extended to out-of-sample forecasting. Their method models a time-varying probability density function nonparametrically using kernel density estimation and schemes for observation weighting that are derived from time series modelling. 
The resulting approach provides directional forecasting signals, specifically for the application of predicting the direction of stock returns. 
Another direction in out-of-sample forecasting is the use of histograms as approximations of the underlying distribution. \citep{ArroyoMate2009} use time series of histograms, with a histogram being available for each observed time point. They propose a kNN-based method to forecast the distribution at a future time point based on the previously observed histograms. However, the method is limited to only being able to forecast an already previously observed histogram, making it more suited for context with recurring patterns. 
Furthermore, motivated by symbolic data analysis \citep{NoirhommeBrito2011}, \citep{DiasBrito2015} proposed an approach that uses linear regression to forecast the density of one variable based on observed histogram data from another variable. However, our objective is an extrapolation of the same variable, but to future time points.
Another direction is the direct modelling of the probability density. Motivated by applications in energy markets, several approaches have been proposed for forecasting energy supply (e.g., wind power) and demand (power consumption). 
\citep{BessaEtal2012} developed a kernel density estimation model based on the Nadaraya-Watson estimator in the context of wind power forecasting. The employed kernels include the beta and gamma kernels as well as the von Mises distribution, underlining the very specialised nature of the approach for use in wind power forecasting. The work of \citep{HeLi2018} is also targeted towards use in the context of wind power forecasting, for which they propose a hybrid model consisting of a quantile regression neural network and a Epanechnikov kernel density estimator. Quantile regression is also used in the work \citep{MokilaneEtal2018} to predict the electricity demand in south Africa for the purpose of long-term planning, while the work of \citep{BikcoraVerheijenWeiland2015} combines an ARMAX and a GARCH model to forecast the density of electricity load in the context of smart loading electric vehicles.
However, the approaches to modelling the probability density presented above all share a strong specificity to their application. %

Density extrapolation and forecasting are part of the more general topic of handling non-stationarity in streaming, time-evolving data.
This problem has gained particular attention in the data stream mining community, where changes in the distribution are commonly denoted as \emph{concept drift} by \citep{WidmerKubat1996} or \emph{population drift} by \citep{KellyHandAdams1999}. 
As discussed in the taxonomy of drift given in \citep{WebbEtal2016}, the drifting subject might be the distribution of features $X$ conditioned on the class label $Y$. Such drift in the class-conditional feature distribution $P(X|Y)$ might result in drift of the posterior distribution $P(Y|X)$. Of particular relevance for our work is \emph{gradual} drift of $P(X|Y)$ and $P(Y|X)$, where the distribution slowly and continuously changes over time, as opposed sudden shift, where it changes abruptly.
Many drift-adaptation techniques have been proposed that aim to incorporate the distribution of the most recently observed labelled data $(X,Y)$ into a machine learning model, as surveyed in \citep{GamaEtal2014}.
However, a challenge in some applications is that no such recent data is available for model update \citep{KremplZliobaiteEtal2014}. For example, in stream classification true labels $Y$ might arrive only with a considerable delay (so-called \emph{verification latency} \citep{MarrsHickeyBlack2010} or label delay \citep{PlasseAdams2016}), or might only be available during the initial training \citep{DyerCapoPolikar2014}. As discussed in \citep{HoferKrempl2012A}, this requires adaptation mechanisms that use the limited available data, which is either recent but unlabelled, or labelled but old. 

These adaptation mechanisms build on \emph{drift models} \citep{KremplHofer2011}, which model the gradual drift over time in the posterior distribution $P(Y|X)$ or the class-conditional feature distribution $P(X|Y)$. Then, they use this for temporal transfer learning from previous time points (source domains) to current or future time points (target domains). This is part of the broader \emph{change mining} paradigm, introduced by \citep{BottcherHoppnerSpiliopoulou2008}. This paradigm aims to understand the changes in a time-evolving distribution, and to use this to describe and predict changes. %
Various drift models and mechanisms for adaptation under verification latency have been proposed. However, they all model the class-conditional distribution of instances, for example by employing clustering assumptions \citep{KremplHofer2011,DyerCapoPolikar2014,SouzaEtal2015}, or calculating changes of a prior \citep{HoferKrempl2012A} or weights \citep{Tasche2014}, or by matching labelled and unlabelled instances \citep{Krempl2011a,Hofer2015,CourtyFlamaryTuia2014}, or they directly adapt the classifier \citep{PlasseAdams2016}.
Thus, they are not applicable to model the changes in a non-parametric, multimodal distribution of unlabelled data, as it is the objective of this work. Thus, the approach proposed in this work complements the existing change mining literature. 
By providing a method for extrapolating $P(X|Y)$, it complements the ex-post drift analysis in \citep{WebbEtal2017} and addresses the calls for better understanding of drift. This might be useful to assess so-called predictability of drift \citep{WebbEtal2016}, and to adapt a classification model in presence of concept drift and verification latency.

\section{Method}\label{sec:method}
\begin{table}\label{tab:notation}

\begin{tabular}{llll}
        $X$ & Observed sample & $R$ & Order of polynomial \\
        $x_i$ & i-th observation in sample & $D$ & Dimensionality of feature space  \\ 
        $T$ & Time attribute of observed sample & $ \gamma_j$ & Weight of j-th basis function\\
        $\tau_i$ & Time value of i-th observation &$ \bm{\gamma}$ & Vector of all $M$ basis weights \\
        $N$ & Observed instance sample size & $B$ & Matrix of regression coefficients\\
        $M$ & Number of basis functions & $\phi_j$ & j-th basis function \\
        $h$ & Bandwidth & $\bm{\phi} $ & Vector of all $M$ basis function \\     
\end{tabular}\\

\caption{Overview of the notation used}
\end{table}
\subsection{Basic Model}
Let $\mathcal{X}$ be a univariate feature for which a sample $X=\{x_1,\cdots,x_N\}$ of size $N$ with associated time values $T=\{\tau_1,\cdots,\tau_N\}$ with $\tau_i \in [0,1]\: \mathrm{for}\: i=1,\cdots,N$  is observed. This forms a data stream segment. 
Since a data stream is a dynamic setting, it is possible that the distribution of $X$ changes as time progresses. This results in the probability density of $X$ at a given time point $t$ to be unequal to that at a future time $t + 1$. This change, referred to as concept drift, is assumed to not being limited to the observed sample, potentially continuing with time. \\
To begin modelling such a possibly changing feature, a model of the density is required. In this work the density is assumed to have the form of an expansion of $M$ normed basis functions $\phi_j(x)$ with $j\in\{1,\cdots,M\}$ and $\int \phi_j(x) dx=1$. The basis functions $\phi_j(x)$ have been chosen as Gaussian density functions located at evenly spaced positions $\mu_j$ throughout the range of $X$, with standard deviation $\sigma=h$ and $h$ being a fixed bandwidth. Other choices of basis functions and placement strategies are also possible. The basis functions could for example also be placed at the location of instances in the observed sample $X$. 
Associated with these basis functions are the weights $\gamma_j$, with $\gamma_j \in [0,1] \:\forall\: j$ and $\sum_{j=1}^{M}\gamma_j = 1$. 
The density then takes the form 

\begin{equation}\label{eq:basisexpansion_static}
f(x)=\sum_{j=1}^M\gamma_j\phi_j(x)=\bm{\phi}(x)^\intercal\bm{\gamma}
\end{equation}

with $\bm{\phi}(x)= (\phi_1(x),\cdots,\phi_M(x))^\intercal$ and $\bm{\gamma}=(\gamma_1,\ldots,\gamma_M)^\intercal$.\\
This formulation of $f(x)$ is insufficient for the setting of a data stream as it does not account for time and the changes that might occur. Based on the model of $f(x)$ as a basis expansion it is proposed to model drift as time-dependent changes in the basis weights, while the basis functions themselves remain unchanged in location and bandwidth. The static basis weight $\gamma_j$ in Eq. \ref{eq:basisexpansion_static} is substituted with $\gamma_j(t)$, a function of time that models the changing basis weight of the $j$-th basis 

\begin{equation}\label{eq:basisexpansion_dynamic}
f(x|t)=\sum_{j=1}^M\gamma_j(t)\phi_j(x)=\bm{\phi}(x)^\intercal\bm{\gamma}(t).
\end{equation}
To fit this model to observed data, which entails determining the basis weight functions $\gamma_j(t)$ for $j=1,\cdots,M$, the likelihood $\mathcal{L}$ of the model
\begin{equation*}
\mathcal{L}=\prod_{i=1}^N f(x_i|\tau_i)=\prod_{i=1}^N (\bm{\phi}(x_i)^\intercal\bm{\gamma}(\tau_i))
\end{equation*}
is maximised by maximising the log-likelihood
\begin{equation}\label{eq:loglikelihood}
\mathrm{log}\: \mathcal{L}=\sum_{i=1}^N \log(f(x_i|\tau_i))=\sum_{i=1}^N \log(\bm{\phi}(x_i)^\intercal\bm{\gamma}(\tau_i)).
\end{equation}

For this it is necessary to consider the properties of the basis weights mentioned above, which translate into two constraints on the basis weight functions, a sum constraint 
$\sum_{j=1}^{M}\gamma_j(t) = 1 \:\forall\: t $
and an interval constraint 
$\gamma_j(t) \in [0,1] \:\forall\: j \:\forall\: t. $

Such constraints are a defining characteristic of what is referred to as compositional data \citep{Aitchison1982}, i.e. data that describes the composition of a whole of $M$ components. It is useful to approach the modelling of the basis weight functions as a compositional data problem, because among the methods developed for the analysis of such data there is one that enables the elegant incorporation of these constraints into the model - the isometric log-ratio (ilr) transformation \citep{EgozcueEtal2003}. The ilr-transformation was proposed by Egozcue et al. to transform compositional data from its original feature space, which for a composition of $M$ components is a $M$-dimensional simplex \citep{Aitchison1982}, to $\mathbb{R}^{M-1}$.

By applying the ilr-transformation to the basis weight functions we acquire $\bm{v}(t)$, their representation in $\mathbb{R}^{M-1}$
\begin{equation*}
ilr(\bm{\gamma}(t))=\bm{U}^\intercal\log(\bm{\gamma}(t))=\bm{v}(t)
\end{equation*}
where $\bm{U}$ is a $M \times M-1$ matrix that is defined as $\bm{U}=\widetilde{\bm{U}}\cdot \bm{D}_2$ with

\begin{align*}
\widetilde{\bm{U}}&=\begin{pmatrix*}[r]
-1&-1&-1&\cdots&-1\\
1&-1&-1&\cdots&-1\\
0&2&-1&\cdots&-1\\
0&0&3&\cdots&-1\\
\vdots&\vdots&\vdots&\ddots&\vdots\\
0&0&0&\cdots&M-1\\
\end{pmatrix*},\:
    \bm{D}_2&=
\begin{pmatrix}
\frac{1}{\Vert\widetilde{\bm{u}}_1\Vert_2}&0&0&\cdots&0\\
0&\frac{1}{\Vert\widetilde{\bm{u}}_2\Vert_2}&0&\cdots&0\\
\vdots&\vdots&\ddots&\cdots&\vdots\\
0&0&0&\cdots&\frac{1}{\Vert\widetilde{\bm{u}}_{M-1}\Vert_2}
\end{pmatrix}\\
\widetilde{\bm{U}}&=(\widetilde{\bm{u}}_1,\ldots,\widetilde{\bm{u}}_{M-1}).
\end{align*}

Due to this transformation we arrive at $\bm{v}(t) = (v_1(t),\cdots,v_{M-1}(t))^\intercal$ , so $M-1$ functions that model the basis function weights in an unconstrained fashion. In this article it is assumed that these functions follow a polynomial regression model of order $R$ 
\begin{equation*}
v_j(t)=\sum_{k=0}^R b_{j,k} \: t^k\qquad j=1,\ldots,M-1.
\end{equation*}
This allows the modelling of $\bm{v}(t)$ as a multivariate regression model
\begin{equation}\label{eq:multivar_regression}
\bm{v}(t) = \bm{B} \bm{a}(t)
\end{equation}
with 
\begin{align*}
\bm{B} =
\begin{pmatrix*} 
b_{1,0} & b_{1,1} & \cdots & b_{1,R+1}\\
\vdots & \vdots & \vdots & \vdots \\
b_{M-1,0} & b_{M-1,1} & \cdots & b_{M-1,R+1}\\
\end{pmatrix*}, \quad
\bm{a}(t)=(t^0,t^1,\ldots,t^R)^\intercal.
\end{align*}

Incorporating Eq. \ref{eq:multivar_regression} into the log-likelihood function in Eq. \ref{eq:loglikelihood} requires the inversion of the ilr-transformation
\begin{equation}\label{eq:backtransform}
\bm{\gamma}(t)=ilr^{-1} (\bm { v}(t))=\frac{\exp(\bm{U}\bm{B}\bm{a}(t))}{\bm{1}'\exp(\bm{U}\bm{B}\bm{a}(t))},
\end{equation}
which is plugged into Eq. \ref{eq:loglikelihood} to arrive at

\begin{align}
\log\:\mathcal{L}(\bm{B}| X,T)&=\sum_{i=1}^N \log\left(\bm{\phi}(x_i)^\intercal \frac{\exp(\bm{U}\bm{B}\bm{a}(\tau_i))}{\bm{1}^\intercal\exp(\bm{U}\bm{B}\bm{a}(\tau_i))}\right)\\
&=\sum_{i=1}^N\log(\bm{\phi}(x_i)^\intercal\exp(\bm{U}\bm{B}\bm{a}(\tau_i)))-\sum_{i=1}^N\log(\bm{1}^\intercal\exp(\bm{U}\bm{B}\bm{a}(\tau_i))\label{eq:loglikelihood_new}
\end{align}

with $\bm{1}$ being a vector of length $M$, each entry being 1. \\

\subsection{Extension: Instance Weighting \& Regularisation}
In this form a maximum likelihood estimation would result in fitting the centre of the training time window best. Since our objective is the extrapolation to future time points outside the training window, it is reasonable to promote a better fit to the data at the end of the training window, i.e. to the most recent data. 
To this end a time-based instance weighting is introduced to the above formulation in the form of multiplying with a weight vector $\bm{w}=(w_1,\cdots,w_N)^\intercal$ that assigns a temporal weight to each observation based on its age. The age of an observation is considered the difference between the largest time value in the observed sample $\max(T)$, i.e. the  most recent one, and the time value $\tau_i$ of the $i$-th observation. The weight assigned to the $i$-th observation is then defined as
\begin{equation}\label{eq:temporalweight}
w_i = \exp\left(\frac{\log(0.5)}{\kappa} \circ (\max(T) - \tau_i)\right)
\end{equation}
where  $\circ$ denotes the Hadamard product, and $\kappa$ is the age of an observation at which a weight of 0.5 is assigned. We recommend to set this value to half of the training window, i.e., $\kappa=0.1$ in our experiments, but depending on the length of the time window, other values for $\kappa$ are also possible. 
Including this temporal instance weighting into Eq. \ref{eq:loglikelihood_new} we arrive at
\begin{equation}\label{eq:weightedloglikelihood}
\begin{split}
\log\:\mathcal{L}(\bm{B}| X,T,\bm{w})=&\sum_{i=1}^N w_i\log(\bm{\phi}(x_i)^\intercal\exp(\bm{U}\bm{B}\bm{a}(\tau_i))) \;-\\
&\sum_{i=1}^N w_i \log(\bm{1}^\intercal\exp(\bm{U}\bm{B}\bm{a}(\tau_i))
\end{split}
\end{equation}
which is the log-likelihood of the model on a biased sample.

Another issue that needs to be addressed is that the regression of the basis weights may suffer from overfitting the observed instances. This would compromise the generalisation to new, future data. To address this a regularisation term $\zeta$ is introduced to the above equation. This penalises the size of the coefficients in $\bm{B}$ with exception of the offset. For this a regularisation strength parameter $\lambda$ is included as well as a $R \times R-1$ matrix $\bm{C}$ so

\begin{equation}\label{eq:regularization}
\zeta =  \lambda \ tr(\bm{C}^\intercal \bm{B}^\intercal \bm{B} \bm{C}),
\quad\bm{C} = \begin{pmatrix}
0&0&\cdots&0\\
1 & 0&\cdots &0\\
0&1&\cdots&0\\
\vdots&\vdots&\ddots&\vdots\\
0&0&\cdots&1
\end{pmatrix},
\end{equation}
which is equivalent to the sum of squares of the coefficient matrix and a common form of regularisation.

\subsection{Optimisation Problem \& Density Model}\label{sec:optprob}

Finally the objective function of the optimisation problem to be solved can be acquired by combining the log-likelihood function including instance weighting as seen in Eq. \ref{eq:weightedloglikelihood} with the regularisation term in Eq. \ref{eq:regularization}. The maximum likelihood estimate of the coefficient matrix $\bm{B}$ is then acquired by solving the unconstrained optimisation problem
\begin{align*}
\argmax_{\bm{B}}\: \log\,\mathcal{L}(\bm{B}| X,T) - \zeta.
\end{align*}

The objective gradient $\nabla l$ is supplied as
\begin{equation*}
\nabla l =( \sum_{i=1}^{N} w_i  (\frac{1}{\beta} \bm{\phi}(x_i) \bm{D} \bm{J})  - \sum_{i=1}^{N} w_i  (\frac{1}{\beta} \bm{1} \bm{D} \bm{J} )) ) - \lambda \ \bm{B} \bm{C} \bm{C}^\intercal, 
\end{equation*}

with  $\bm{J}=\frac{\partial\: \bm{U}\bm{B}\bm{a}(\tau_i)}{\partial \bm{B}}$ and $\bm{D}$ being a diagonal matrix with $\frac{\partial \:exp(\bm{U}\bm{B}\bm{a}(\tau_i))}{\partial \bm{B}}$ on the diagonal. The detailed derivation of the gradient can be found in the appendix.\\

To solve the optimisation problem we use MATLABs Global Optimization Toolbox and its implementation of the Quasi-Newton algorithm as provided by the function \textit{fminunc}. Since only functionality for minimisation problems is provided, we redefine the optimisation problem stated above to 
$$\argmin_{\bm{B}}\: -\left(\log\,\mathcal{L}(\bm{B}| X,T) - \zeta\right)$$ and use $-\nabla l$ accordingly. Besides enabling the use of the supplied gradient and using a larger optimality tolerance of $1e^{-4}$ (instead of default $1e^{-6}$) we use the functions default parameters. \\
This solver configuration was used in a multiple starting point search executed via the \textit{MultiStart} function provided by the previously mentioned toolbox. For this the 'artificial bound' parameter used for starting point generation is set to 2 and the number of start points was set to 4, the choice of the latter is a trade-off between higher optimality of the solution and shorter computational runtime.

After the optimisation problem is solved the coefficients in $\bm{B}$ can be used to extrapolate the density of the model. For this the expression in Eq. \ref{eq:backtransform} is plugged into Eq. \ref{eq:basisexpansion_dynamic}, giving the models density estimate for a sample $(x,t)$ as

\begin{equation}\label{eq:finaldensity}
f(x|t)=\bm{\phi}(x)^\intercal \frac{\exp(\bm{U}\bm{B}\bm{a}(t))}{\bm{1}^\intercal\exp(\bm{U}\bm{B}\bm{a}(t))}
\end{equation}

\section{Experimental Evaluation} \label{sec:expeval}

The goals of the experimental evaluation are twofold. First, to assess the quality of the densities predicted by the models for future time points. This is done by comparing the proposed method to the other two methods that are available in the literature for this problem. Second, to investigate the behaviour of the proposed method in the form of an analysis of its sensitivity with respect to the model hyperparameters and its computational runtime. 
The experiments are conducted in the 2017b version of MATLAB with the exception of the EDD method, for which an implementation in Python 2.7 has graciously been provided by the author of the EDD method. All experiments are conducted on the 'Gemini' cluster of Utrecht University, using a PowerEdge R730 with 32 HT cores and 256 GB memory. \\
For the experiments a range of data sets have been selected, in part real and artificial, which show various different kinds of drift over time. These data sets will be discussed in detail in the following. 

\subsection{Data}
Four different artificial data sets are generated to simulate different drift patterns. The use of artificial data in this context has the advantages that both the drift pattern is explicitly known as well as the data generating process in general, so the models' estimates can be compared to the true density. \\
All four artificial data sets are generated by a mixture of three or four skew-normal distributions. To simulate different kinds of drift certain parameters of the mixture distribution are subjected to gradual changes over time, which is reflected in the names of the data sets:
\begin{itemize}
\item \emph{meandrift} - four components with the location parameter of all but the first component changing over time.
\item \emph{weightdrift} - three components, mixture weight of the second decreases over time, weights of the other two increase.
\item \emph{sigmachange} - three components with scale parameter changing over time.
\item \emph{staticskewnormals} - four unchanging components.
\end{itemize}
All artificial data sets contain 25000 instances in the range of $[0,12]$ which are equally distributed over 120 unique time points in the interval $[0,1]$. 

The first real-world data set used in the experiments is taken from the website of the US peer-lending company 'Lending Club'\footnote{\url{https://www.lendingclub.com/}}, containing the accepted 
loan requests of the years 2007-2017. This data was then preprocessed by applying to each variable separately a Box-Cox transformation \citep{BoxCox1964}, as implemented in the R package MASS \citep{VenablesRipley2002}.
Since the proposed method in this form can only handle univariate data the processed lending club data set is split. This results in each feature forming a separate data set, which is named after the feature. Each of these new data sets contains the feature values as well as the time stamps. In total the data contains 120 different time stamps, corresponding to the monthly data of 10 years.
The features that have been selected from the original 75 are:
\begin{itemize}
\item \emph{dti} - a ratio calculated using the borrower’s total monthly debt payments on the total debt obligations, excluding mortgage and the requested LC loan, divided by the borrower’s self-reported monthly income.
\item \emph{int\_rate} - interest rate on the loan 
\item \emph{loan\_amnt} - the listed amount of the loan applied for by the borrower. If at some point in time the credit department reduces the loan amount, then it will be reflected in this value.
\item \emph{open\_acc} - the number of open credit lines in the borrower's credit file.
\item \emph{revol\_util} - Revolving line utilisation rate, or the amount of credit the borrower is using relative to all available revolving credit.
\end{itemize}

One notable characteristic of the lending club data sets is the vastly bigger amount of data points at later time points in the stream. The 25th percentile lies at $t=0.7$ and the 75th percentile at $t=0.91$. This is due to the fact that the data set begins not long after the starting of the company and therefore also shows the increased amounts of credit applications.

Another data set that is taken from the real world is what will be referred to here as the 'pollution' data set. It is created from a Kaggle data set about air pollution in Skopje, Macedonia, between the years of 2008 to 2018\footnote{See \url{https://www.kaggle.com/cokastefan/pm10-pollution-data-in-skopje-from-2008-to-2018}}. The original version of the data set contains measurements of carbon monoxide (CO), nitrogen dioxide ($\mathrm{NO}_2$), ozone ($\mathrm{O}_3$) as well as particulate matters smaller than 10 and 2.5 micrometers (PM10, PM25), measured at 7 different measurement stations in the city. While technically the measurements have been taken on an hourly basis within the time frame of 2008 to 2018, there are often missing values for certain stations or certain compounds.
Since the original data represents time series of single measurements, it is not exactly fitting for our purposes, since we are interested in forecasting the distribution of the data. 

To arrive at a more suitable data set the data was split into one subset per feature, resulting in 5 subsets with each a feature and the time variable. The separation by measuring station was removed, meaning that each hourly time stamp had at most 7 measurements associated to it. Each of these subsets underwent the same preprocessing. First, all entries whose feature value were missing have been discarded. Then all entries whose time stamp does not lie within the inclusive interval of 4 pm till 8 pm (considered the time frame of evening rush hour by common definition) were removed to eliminate the intra-day variation in the data and instead focusing on the trends on a monthly basis. To this end the time stamps were generalised to month-level by dropping the day component of the time stamp. Finally the time variable was normalised to a scale from 0 to 1 and the feature values were either box-transformed if the fitted $\lambda$ value was not 0 and log-transformed if equal 0. In the end the processed data set represents the transformed compound measurements during evening rush hour in Skopje on a monthly basis, allowing for an analysis of the change in the distribution over time.

\begin{figure}
\centering
\includegraphics[width=1\textwidth]{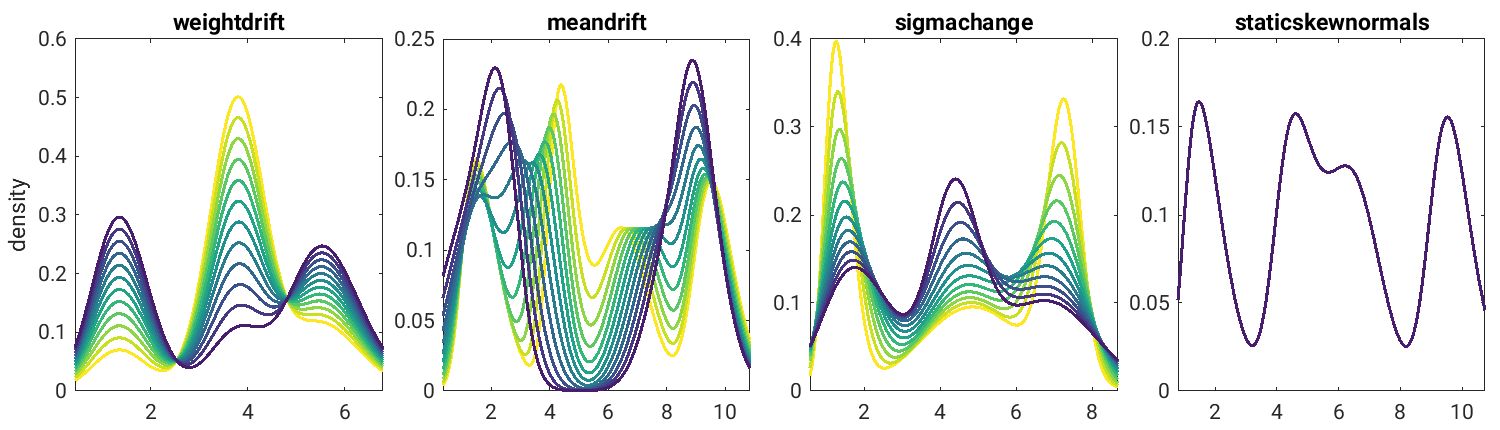} 
\\ 
\includegraphics[width=1\textwidth]{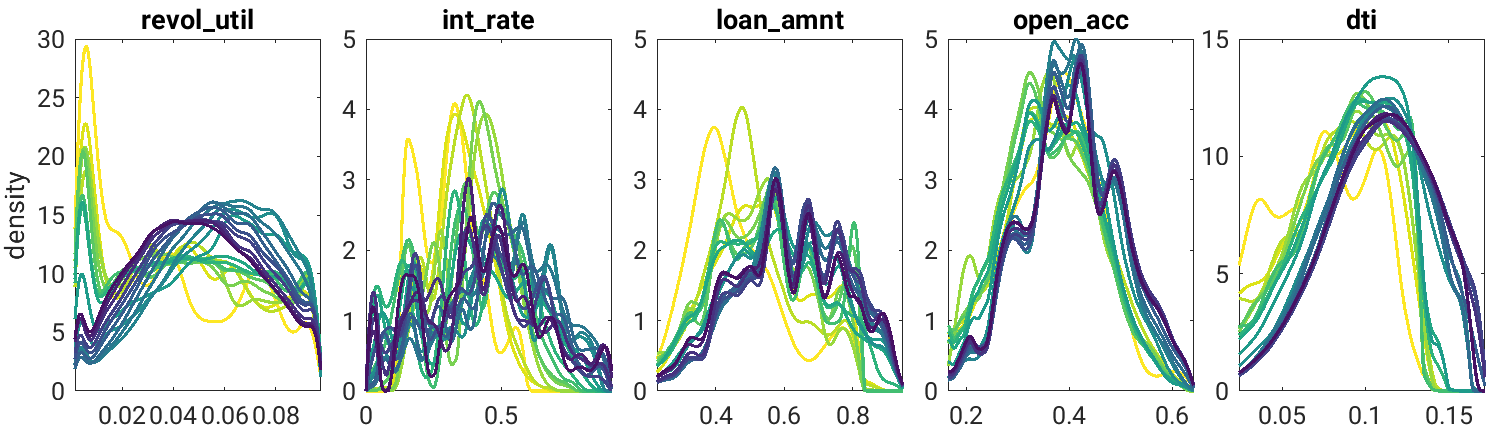} 
\\
\includegraphics[width=1\textwidth]{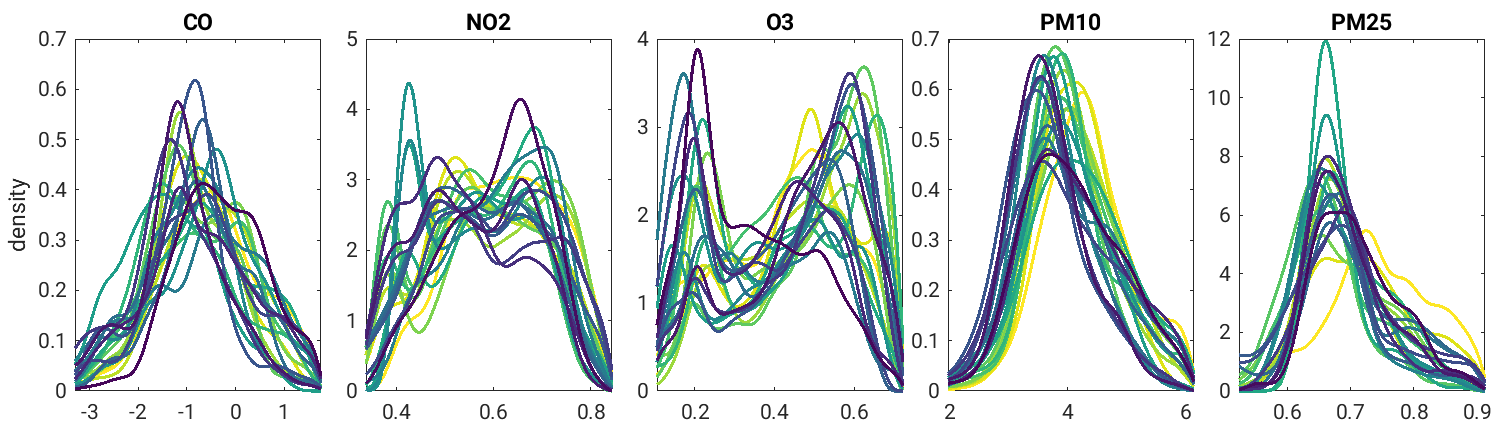} 
\caption{True density of artificial data (top row), baseline density of lending club data (middle row) and baseline density of pollution data (bottom row) illustrated over time. Not all time points have been plotted for visibility's sake. Time progression colour-coded from light (t=0) to dark (t=1).}\label{fig:datadrift}
\end{figure}

Figure \ref{fig:datadrift} illustrates the changes in the density over time on both the real and artificial data sets. The X-axis shows the feature values ($X$), while the Y-axis shows the probability density of $X$. For the artificial data set this is the true density, for the real data it is an approximation as will be discussed later. The shade of the different lines indicates the time point associated to the density curve, with lighter shades representing earlier and darker shades representing later time points. The middle right plot in Fig. \ref{fig:datadrift} shows the change in the approximated density on the lending club 'dti' data set, showing a slow shift to the right that is fairly simple, while the plot of the 'interest rate' data set shows a more complex change . Many small changes at multiple locations make this pattern fairly complex and noisy. In contrast to this, the drift patterns of the artificial data sets are intentionally simple to clearly simulate certain causes for a changing distribution. \\

To give a comparison to the proposed method regarding the quality of the predicted density, two additional methods have been employed. The first of these is a static version of the proposed method, which fits the basis expansion model with static weights instead of modelling a time-dependent weight function. As such, this model does not account for the temporal dimension of the data. The second method used is the "Extrapolating the Distribution Dynamics" (EDD) approach proposed by C.H. Lampert \citep{Lampert2015}. Since EDD is based on kernel density estimates, the number of training instances needs to be kept in mind when tuning the bandwidth parameter of the model. If the number of training instances during model selection is very different from the number of instances in the experiments training window, the kernel bandwidth is likely to be unfit, resulting in poor performance. To account for this it was ensured in the experiments that the training set presented to EDD in the experiments comprises the same amount of instances as there are in the respective model selection training set. The procedure for this is as follows: if the training window of the model selection comprises less than 10\% of the entire data set, then all instances in this window are used; if the model selection training window contains more than 10\% of the overall instances, then a random subsample corresponding to 10\% of the data is used. When training the model during the experiments it is ensured that the number of training instances does not exceed the number of instances used during model selection. If the experiment training window contains more instances than that, a random subsample is used - if it contains less, all samples are used.

\begin{figure}
\centering
\includegraphics[width=0.7\textwidth]{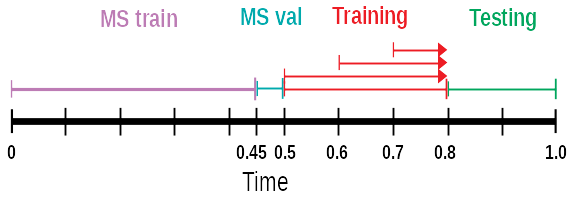} 
\caption{Segmentation of the data for the experiments. Separate time intervals for training as part of the model selection ('MS train') and the validation of its models ('MS val'), as well as for the training of the final models with three different training window sizes ('Training') and their evaluation ('Testing')}
\label{fig:expsetup}
\end{figure}

The experiment setup comprises a model selection phase elaborated on in Sec. \ref{sec:model_selection}, a training phase using the selected hyperparameters and the application of the trained models to predict the density in a previously unseen segment of the data stream. 
Thus, on an infinite stream the model selection is done at the beginning, while training and prediction phases are repeated over time. 
The data sets  have been partitioned into several time windows as illustrated in Fig. \ref{fig:expsetup} for the experiments. Consider that $t=0$ indicates the first time point in the data set and $t=1$ the last time point. In  this setting, we consider the data until $t=0.8$ as available historic data and the data after that as entirely unseen.
We chose the time frame $[0.0, 0.5]$ for model selection, because of the strongly skewed distribution of instances over time in the lending club data set. This way there is a sufficiently large initial sample for all algorithms.
Therein, the interval $[0.0, 0.45[$ is used to train models with different hyperparameter combinations and the interval $[0.45, 0.5[$ is used to evaluate them. The segment of $[0.5, 0.8[$ was used to train the models with the previously selected hyperparameter combinations. In order to investigate the influence of the training window size on the model performance, the methods were trained once on each of the three sub-intervals $[0.5, 0.8[$, $[0.6, 0.8[$ and $[0.7, 0.8[$. These 3 training windows per method result in 9 final models per data set that are finally evaluated.\\
Each of the trained models is then applied to predict the out of time density, that is at the time points within the unseen testing window. This testing window spans the remaining interval of $[0.8, 1.0]$ and is the same for all models. In the evaluation we will refer to the time difference between the end of the training window and the time point of the prediction as "latency". 

The quality of a model's density estimate for a given time point is measured by the mean absolute error (MAE) of the prediction and the true density, evaluated at a previously defined set of points. This error measure was chosen because it is commonly used in forecasting and provides a straight-forward interpretation. Computing the error measure on the artificial data sets is simple, since the data generating process is entirely known and the true density can be computed. This however proves problematic with the real-world data, where the underlying process that generated the data is unknown and only random samples of varying size drawn from this process are available. To still evaluate the quality of the predicted density in this case, it is required to approximate the density based on the observed samples. 

To approximate the true density at a time point $t_i$ we consider the instances within a time window of size 4 around that time point to smooth potential noise, i.e. $\{t_{i-4},\cdots,t_i,\cdots,t_{i+4}\}$. Based on this set of instances of size $S$ an ensemble of 9 smoothed histograms is created, with each histogram using a different bin size $b$. The set of used bin sizes consists of 8 incremental integer steps, 4 in positive and 4 in negative direction, around the result of Sturges' formula $b_s = (\log_2 S) +1$ \citep{Sturges1926}. For each of these histograms we then compute the relative number of samples per bin with respect to the number of samples and divide it by the bin size. This density scaled, relative frequency is then associated to the bin centres and used to fit a cubic spline. Each smoothed histogram then approximates the density by the splines function value where it is greater than zero, while the density is regarded as equal to zero where the splines function value is smaller or equal zero. Each spline in the ensemble is then evaluated over the same set of points across the domain of $X$. The mean of the resulting 9 vectors then forms the approximation of the true density. This approximation is in the following only referred to as 'baseline density'.

\subsection{Model Selection}\label{sec:model_selection}

Both the method proposed in this article as well as EDD \citep{Lampert2015} require hyperparameters that need to be determined as part of the model selection step in the experiments. The proposed method requires four hyperparameters, namely the number of basis functions $M$, their bandwidth $h$, the order of the polynomial $R$ in the multivariate regression and the regularisation factor $\lambda$. EDD requires the Gaussian kernel bandwidth $\sigma$ and also the regularisation factor $\lambda$. 

In order to determine these hyperparameters a model selection step is incorporated into the experiment design, using the data designated for the model selection phase as discussed earlier.
For the $\lambda$ hyperparameter of EDD the author used $\frac{1}{N}$ as a default value for his experiments on the artificial data sets. We based the parameter search space for $\lambda$ on this suggestion by multiplying it by a scaling factor $b \in \{0, 0.25, 0.5, 0.75, 1.0, 1.25, 1.5\}$.
The parameter search space for $\sigma$ consists of 20 evenly spaced numbers in the interval $[0.001, 0.3]$. For each possible combination of values in these two parameter search spaces an EDD model is trained and the parameter values of the model that scored the lowest MAE on the validation data is selected for use in the experiments for the given data set.

The model selection process for the proposed method is divided into two phases. First, $M$ and $h$ are optimised, since they determine the general fit of the basis expansion model. For $M$ the values in $\{10,12,14\}$ are considered. Since $M$ basis functions are spaced equidistantly and are fixed in their location, the coverage of the feature space by the basis functions is influenced by both $M$ and $h$ - e.g. a particular value for $h$ might be a good choice for $M=14$, but a poor choice for $M=10$ for example by resulting in gaps between the bases. Because of this connection the limits of the searched interval $[h_{min},h_{max}]$ are computed depending on $M$ and the range of the available data as 
\begin{align*}
h_{min}=\frac{P_{0.99}(X_{mst})-P_{0.01}(X_{mst})}{M} \cdot 0.5\\
h_{max}=\frac{P_{0.99}(X_{mst})-P_{0.01}(X_{mst})}{M} \cdot 1.2
\end{align*}
with $P_z(\cdot)$ denoting the z-th percentile function and $X_{mst}$ the data within the model selection training window. 

Again, the hyperparameter combination that yields the smallest MAE is selected. This is then used in the second phase of the model selection in which $R$ and $\lambda$ are determined via a linear search of the parameter space for both parameters, with $R \in \{1,2,3\}$ and $\lambda \in \{1,2,3,4,5\}$. Note that the search spaces for these parameters are based on experience with experiments prior to those presented here. A detailed account of the models sensitivity to different hyperparameter values will be presented in Sec. \ref{sec:SA}.

\subsection{Results}\label{sec:results}
In this section we will first address the question of the models predictive quality, for which the three groups of data sets (artificial, lending club and pollution) will be regarded sequentially. Then the results of a series of experiments with the goal of investigating the sensitivity of the proposed method with respect to its hyperparameters will be presented. Finally the computational effort of the involved methods will be discussed.

\subsubsection{Artificial Data}\label{sec:results_arti}
\begin{figure}
    \centering
    \begin{subfigure}[b]{0.48\textwidth}
        \includegraphics[width=\textwidth]{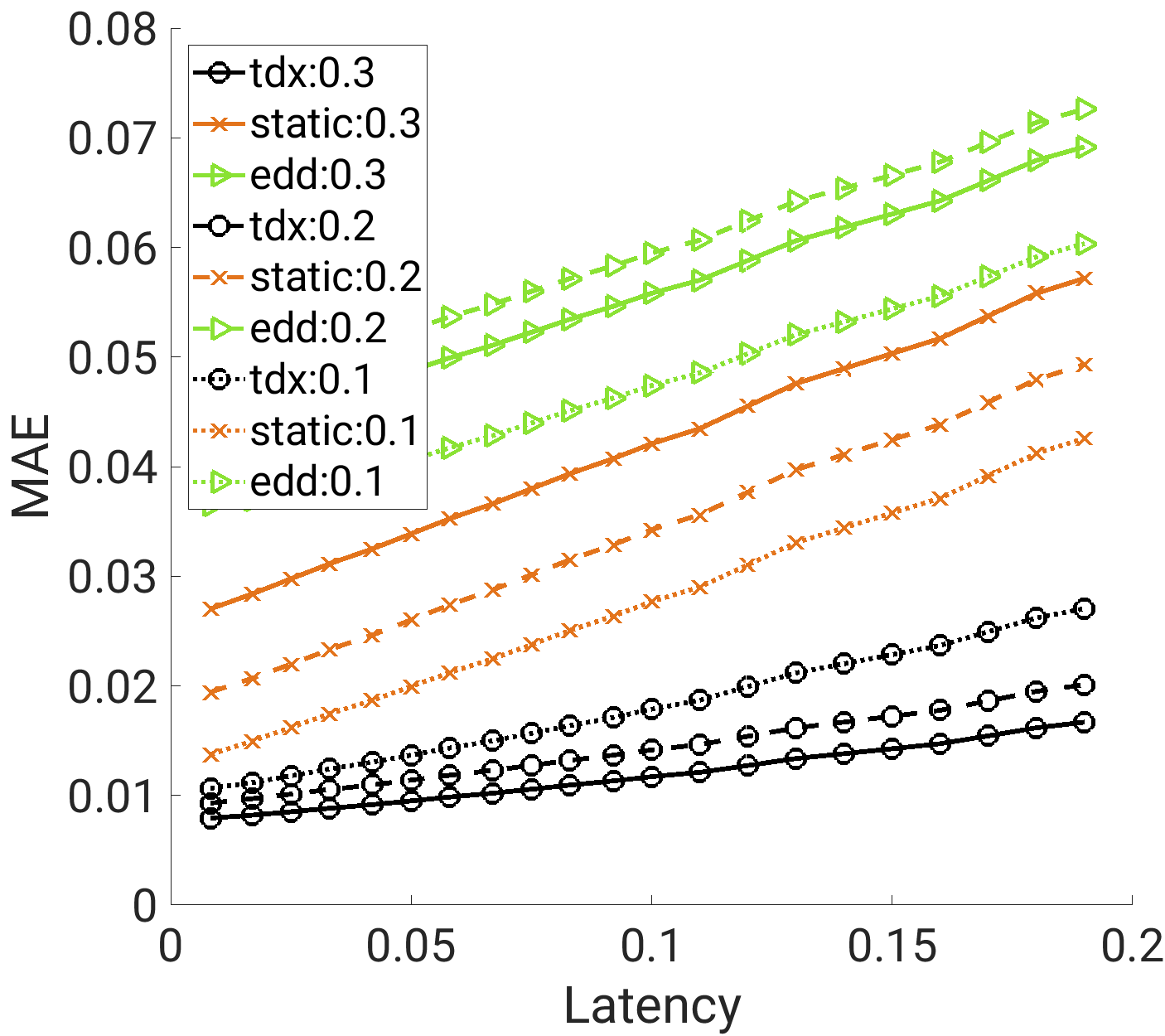} 
        \caption{Weightdrift}
        \label{fig:error_weightdrift}
    \end{subfigure}
    ~ %
    \begin{subfigure}[b]{0.48\textwidth}
        \includegraphics[width=\textwidth]{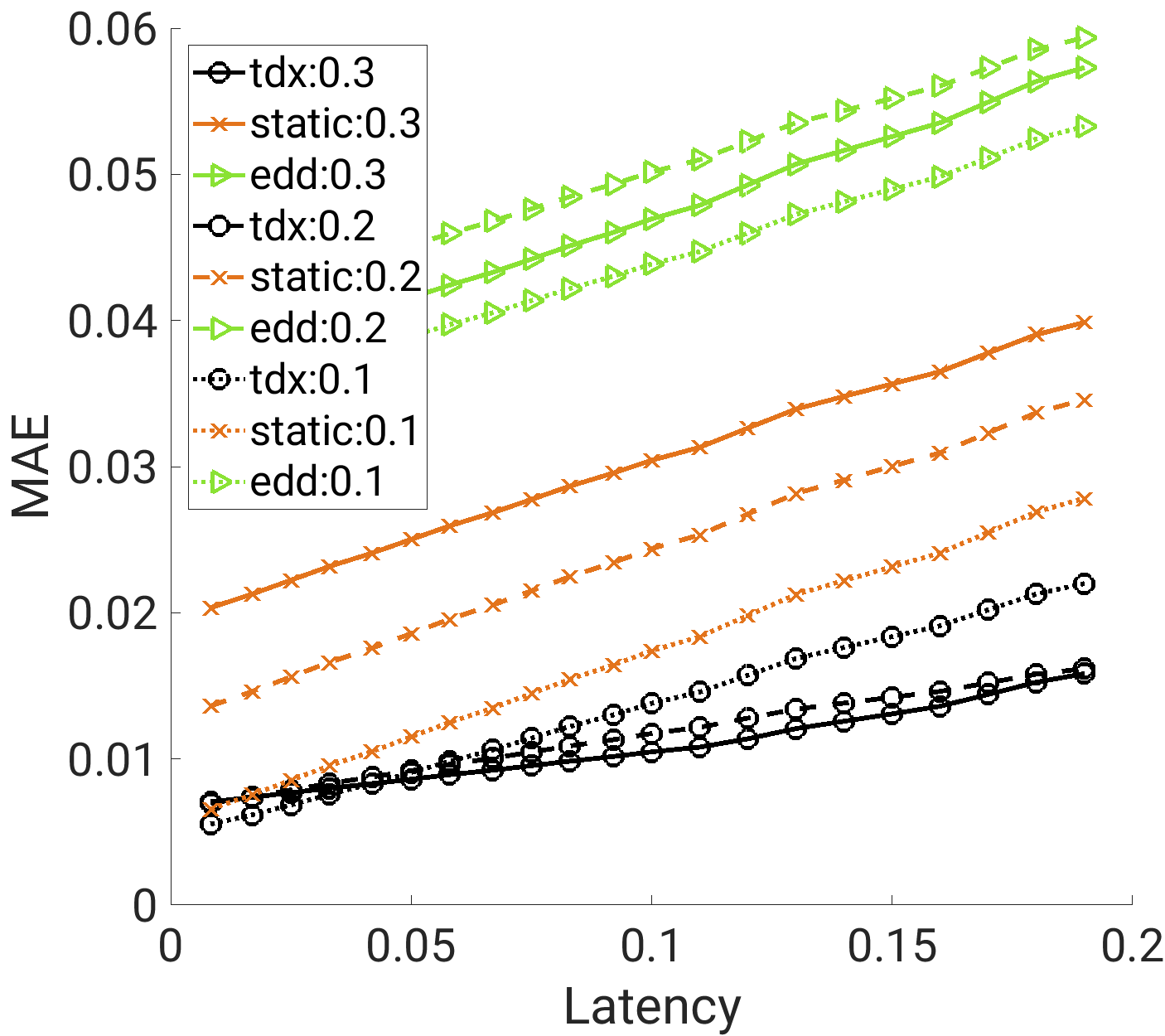} 
        \caption{Meandrift}
        \label{fig:error_meandrift}
    \end{subfigure}
    
    \begin{subfigure}[b]{0.48\textwidth}
        \includegraphics[width=\textwidth]{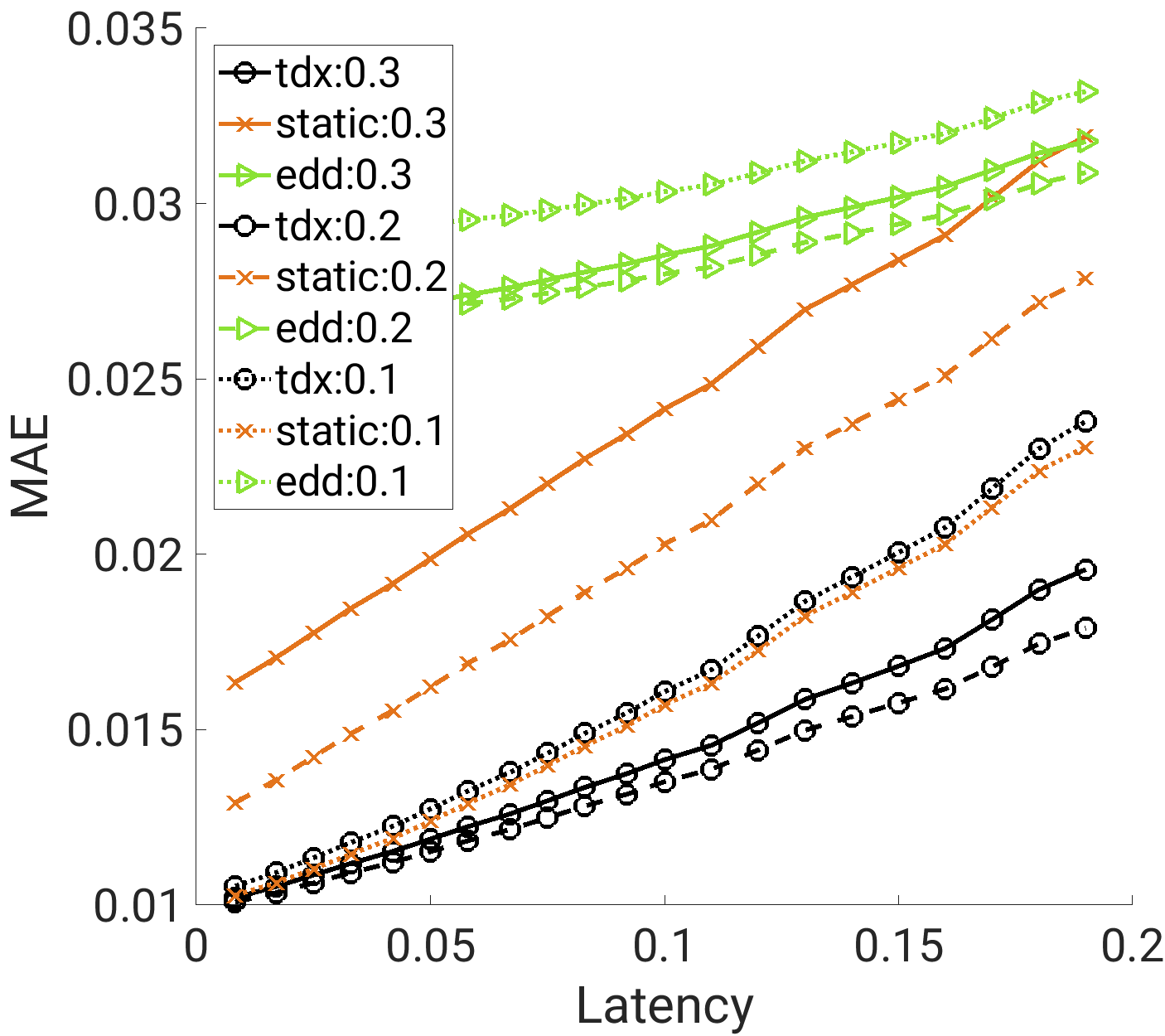} 
        \caption{Sigmachange}
        \label{fig:error_sigmachange}
    \end{subfigure}
    ~
    \begin{subfigure}[b]{0.48\textwidth}
      \includegraphics[width=\textwidth]{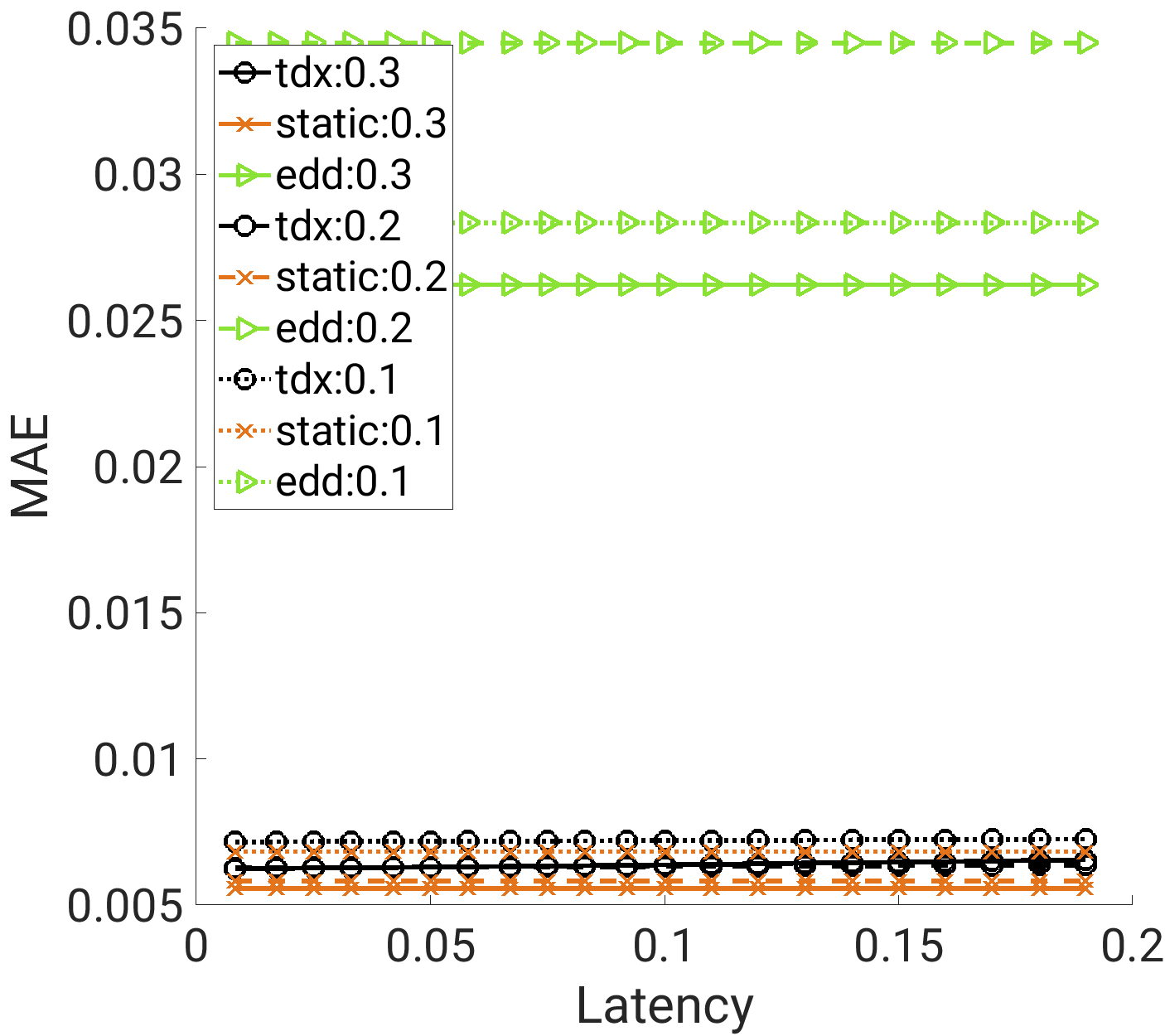} 
      \caption{Staticskewnormals}
      \label{fig:error_staticskewnormals}
    \end{subfigure}
    \caption{MAE of each method (indicated by colour) for each training window length (indicated by linestyle) for different latency values on the artificial data sets.}\label{fig:error_artidata}
\end{figure}

The artificial data sets provide an important starting point for the evaluation of the proposed method since the drift patterns are known and clearly defined. This way one can evaluate whether a particular kind of pattern is recognised. \\
The weightdrift pattern matches the assumptions of the TDX model the best, given that the weightdrift data is generated by a mixture distribution whose mixture weights are linearly changing over time. Fig. \ref{fig:error_weightdrift} shows the MAE of the different models (tdx, static, edd) for the three different training window lengths ( 0.1, 0.2, 0.3) indicated by linestyle for a range of latency values (X-axis). As mentioned earlier, the latency is the time difference between the end of the training window and the time point of the forecast. TDX is performing best on this pattern for all three training window sizes and it can be seen that the length of the training window influences the error of the model and its change over time. The TDX model with the longest training window shows both a lower error and a slower increase in error over time compared to the TDX models with window lengths of 0.2 and 0.1 respectively. The opposite effect can be observed with the static model, which performs better with a smaller training window.\\
On the meandrift pattern the location parameters of the sub-distributions in the mixture linearly change over time, which results in a less smooth drift compared to that of weightdrift as seen in Fig. \ref{fig:datadrift}.
Nonetheless TDX also performs best on this pattern as can be seen in Fig. \ref{fig:error_meandrift}. Here the difference between the different training window lengths for TDX is smaller. Although the model with the longest training window performs marginally worse for smaller latency values, it later consistently performs better than both models with shorter training windows. 
The static models show a slightly higher error and increase in error, again with the smallest training window resulting in a lower error.\\
On the sigmachange data the drift is simulated by a linear change in the standard deviation of the components in the mixture distribution, resulting in the density change shown in Fig. \ref{fig:datadrift}. Fig. \ref{fig:error_sigmachange} shows that all TDX models except the one with the shortest training window perform consistently better than the static model, showing lower error and a slower increase in error over time. \\
Finally the performance on the staticskewnormals data set is illustrated in Fig. \ref{fig:error_staticskewnormals}, which does not include any change in the distribution over time at all. Although all TDX and static models are fairly close in terms of error ($<0.0025$) the best TDX model only scores the third lowest error. All TDX models show almost no increase in error over time, indicating that the model was able to recognise the absence of drift.

\begin{figure}
\centering
  \includegraphics[width=0.8\textwidth]{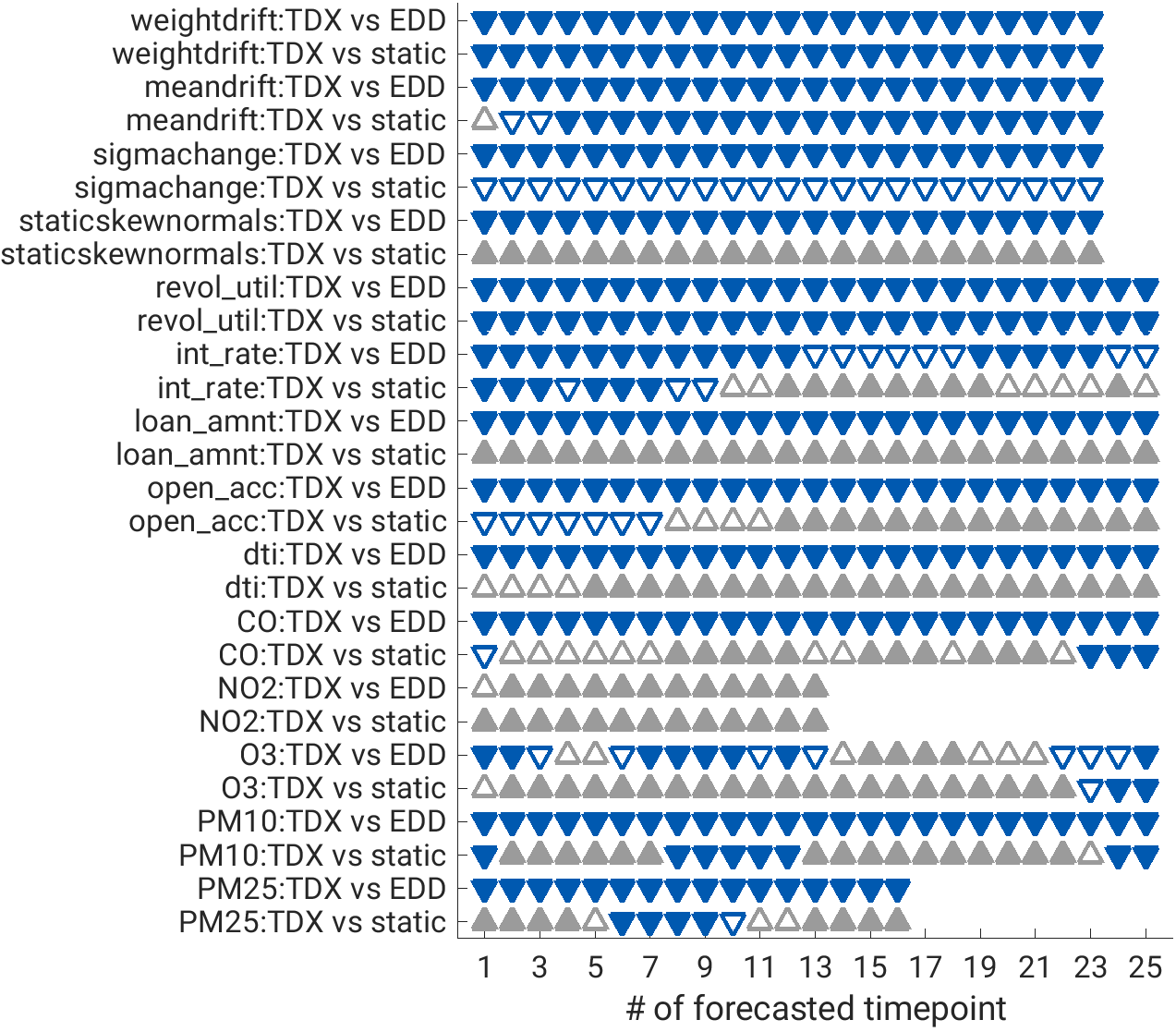} 
  \caption{Summary of TDXs estimation error in comparison to EDD and static. X-axis shows the forecasted timepoint, the markers on the Y-axis indicate TDXs relative error: triangle downward, if MAE of TDX is smaller than that of the other method, triangle upward if MAE of TDX is bigger. Filled out markers indicate that the difference is significant.}
  \label{fig:wilcox}
\end{figure}

To statistically validate these results a series of two-sided Wilcoxon signed-rank tests was performed. For each data set, we selected the best performing model of each method with respect to the training window size based on the summed MAE over all forecasted timepoints. Based on this selection we considered two scenarios, ``TDX vs EDD'' and ``TDX vs static''. For each time point within the forecasting time window, the absolute deviations of the method's predicted density to the true/baseline density was computed at the same 200 equally spaced points within the domain of $X$. These error distributions were then tested with the two-sided Wilcoxon signed-rank test in order to determine whether the differences of these distributions are significant 
at a significance level of 0.01.\\
The results of these tests are illustrated in Fig. \ref{fig:wilcox}. The orientation of the triangular markers indicated whether the MAE of TDX is lower (downward) or larger (upward) than the other method in the test scenario, indicated in the labels of the Y-axis. If the difference in the error distributions at a given timepoint is significant according to the test mentioned above, the marker appears filled, otherwise empty.\\ 
These results show that the proposed method performs significantly better than EDD on all artificial data sets for all forecasted time points. Compared to the static model it also scores consistently lower MAE on weight- and meandrift with exception of the very first time point on meandrift. These results are significant on both data sets over all time points except for the first 3 on meandrift. On the sigmachange data TDX shows a consistently lower MAE than the static model, but the differences are not significant at any of these time points. Finally, although the MAE of TDX's predictions is only slightly larger than that of the static model for all time points on statiskewnormals, the differences are significant for all forecasted timepoints.\\
In summary the proposed method manages to capture the drift patterns on weightdrift, meandrift and sigmachange well and performs better than the static model. On the staticskewnormals data set the TDX models consistently have higher error than the static model, showing that on a entirely static data set the static model is hard to surpass.

\subsubsection{Lending Club Data}
\begin{figure}[!h]
\centering
\begin{minipage}{.48\textwidth}
  \centering
 \includegraphics[width=\linewidth]{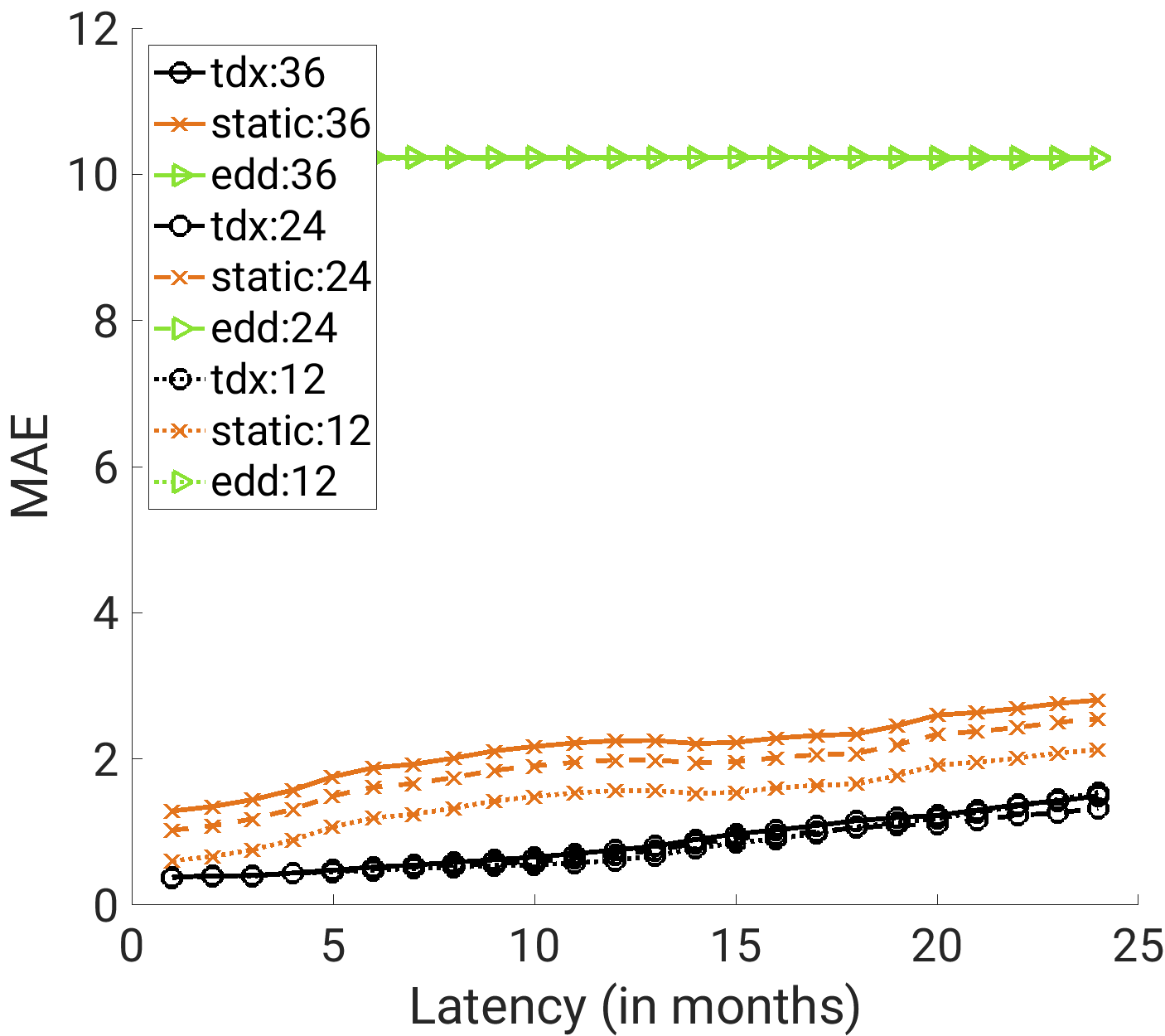} 
  \caption{MAE on revol\_util for different training window sizes and latency values}\label{fig:error_revol_util}
\end{minipage}%
\hfill
\begin{minipage}{.48\textwidth}
  \centering
  \includegraphics[width=\linewidth]{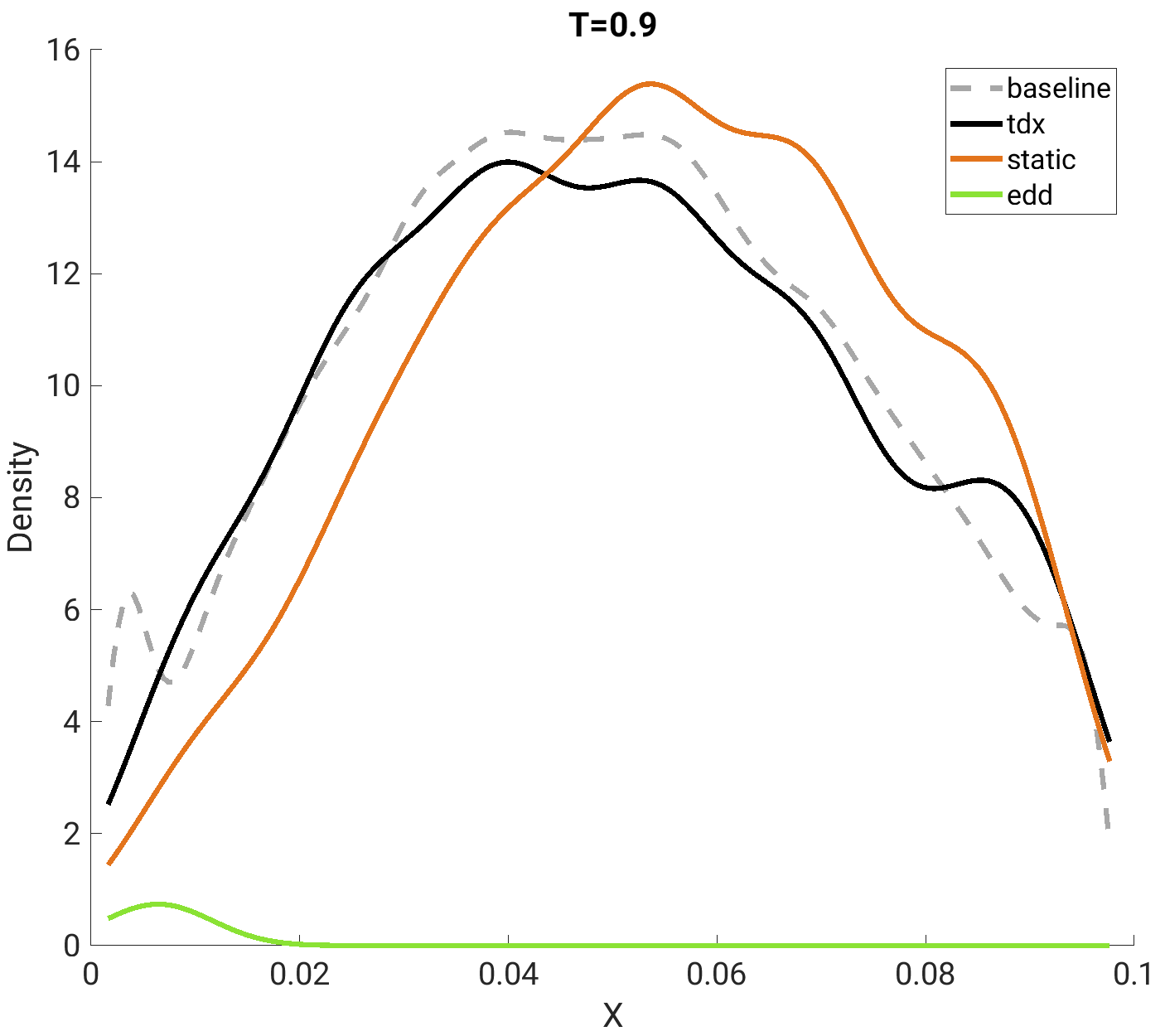} 
  \caption{Baseline and predicted density at T=0.9 (12 months latency) of models with training window length of 12 months on the revol\_util data}\label{fig:revol_util_density_108} 
\end{minipage}
\end{figure}

The lending club data sets are the first set of real-world data sets used in the experiments. Since the data contains 10 years worth of monthly data, the time variable is measured in months on these data sets as give a better understanding of the presented time frames.\\
As the first of these  data sets the revolving line utilisation rate (short 'revol\_util') is considered. A look at the evolution of the baseline density of this data set in Fig. \ref{fig:datadrift} (second row, first from left) shows interesting, non-monotonous drift pattern. A peak on the left edge of the domain of $X$ diminishes as the density shifts to form a new peak around $X\in[0.06,0.08]$ that then shifts left towards $T\in[0.03,0.06]$. The results in Fig. \ref{fig:error_revol_util} show that on the revol\_util data set TDX performs well with all three window sizes resulting in very similar error as well as almost identical, slight increase in error over time. Throughout the entire forecasting period all three TDX models achieve lower errors than the static models, while EDD scores an enormous error due to the bandwidth selection of the grid search being unfit for this later segment of the data set. The evolution of the density on this data set as shown in Fig. \ref{fig:datadrift}  provides a hint as to why a poor choice of bandwidth resulted from the model selection, since the density distribution has a very different shape during the earlier time points. This indicates that EDD's parameters might be very sensitive to distributional changes. Looking at the predicted density as well as the baseline density in Fig. \ref{fig:revol_util_density_108} one notices that even with a latency of 12 months the TDX model manages to anticipate the change in the density, while the forecast of the static model still reflects the density at the end of the training window.\\
Fig. \ref{fig:wilcox} shows that TDX's prediction not only result in a lower MAE than the other two methods, but that the distributions of the absolute error are also significantly different for all forecasted timepoints.

\begin{figure}%
\centering
\begin{minipage}{.48\textwidth}
  \centering
 \includegraphics[width=\linewidth]{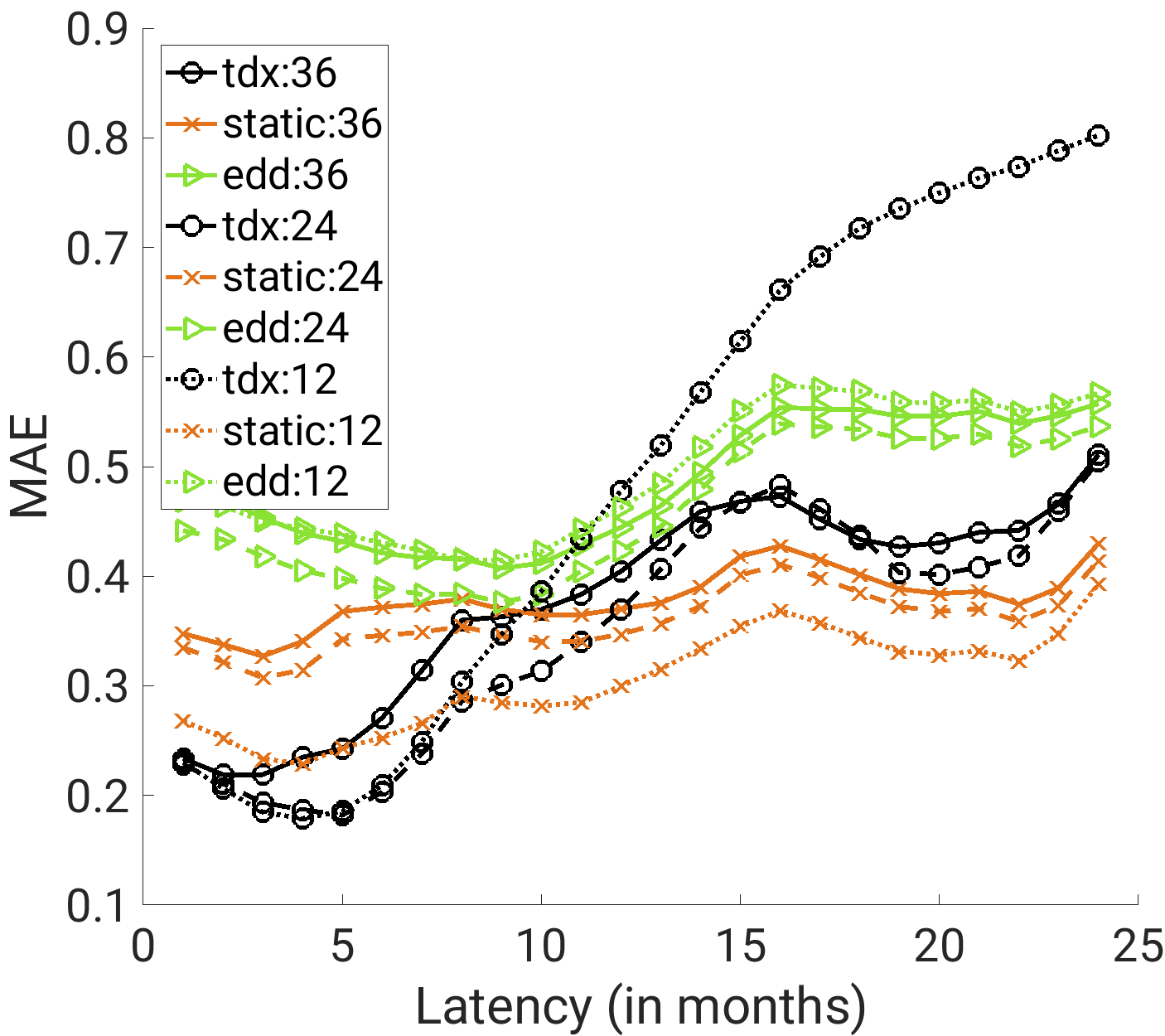} 
  \caption{MAE on int\_rate for different training window sizes and latency values}
  \label{fig:error_int_rate}
\end{minipage}%
\hfill
\begin{minipage}{.48\textwidth}
  \centering
  \includegraphics[width=\linewidth]{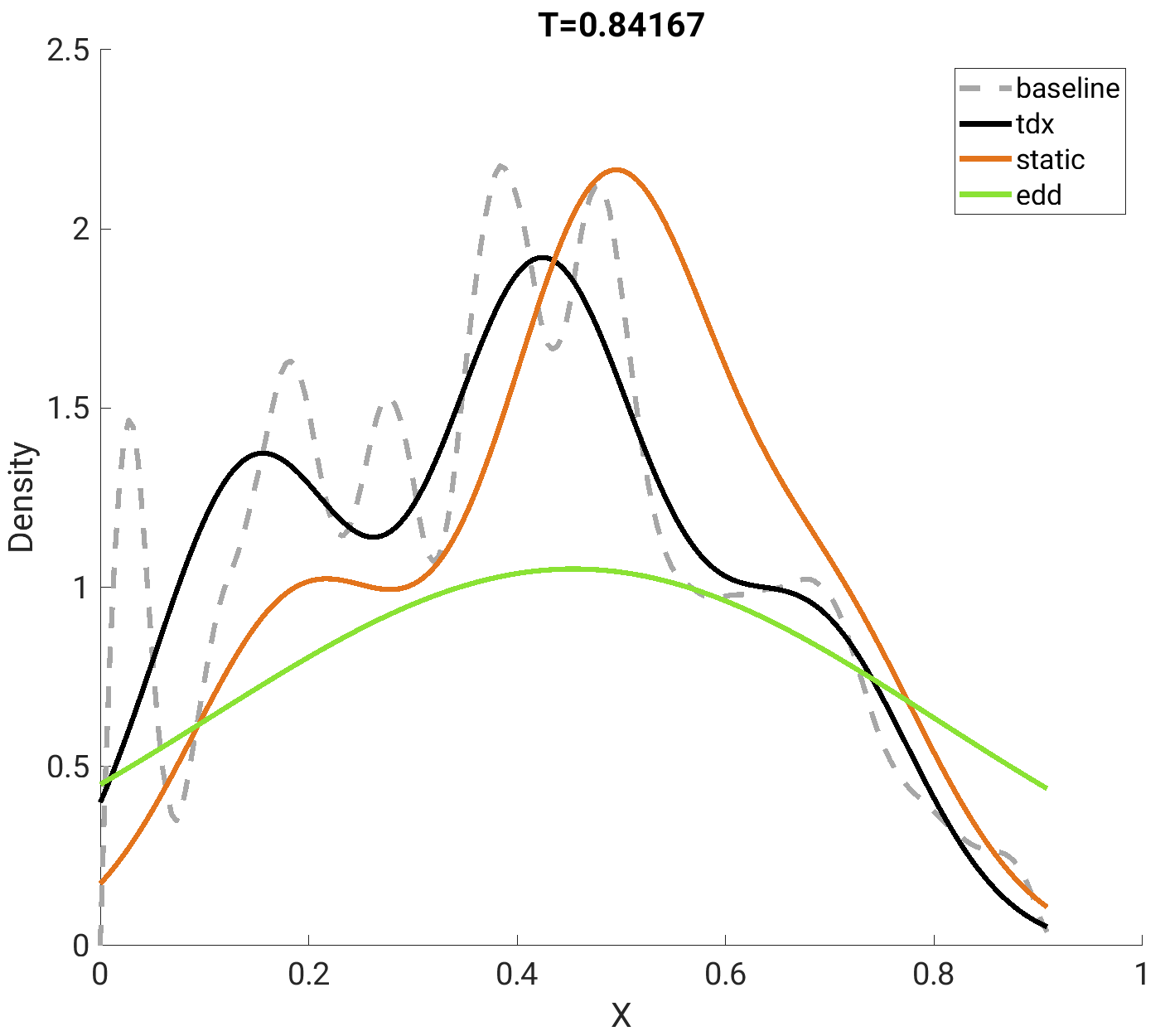} 
  \caption{Baseline and predicted density at T=0.841 (5 months latency) of models with training window length of 24 months on int\_rate} \label{fig:int_rate_density_101}
\end{minipage}
\end{figure}

A more volatile and mobile drift can be observed on the interest rate data set (short 'int\_rate') in Fig \ref{fig:datadrift} (second row, second from left). Multiple smaller movements in the density make a difficult, non-monotonous pattern. Fig. \ref{fig:error_int_rate} shows that within the first 7 months of latency the TDX models with 12 and 24 months of training data reach the lowest and second-lowest error respectively. After this point the static model with a 12 month window surpasses them, remaining the lowest error model for the rest of the forecast period. For the TDX model with a 12 month training window a quick increase in error can be observed for latencies $> 8$ months, which is likely a result of the non-monotonous and often reversing trends within the data. Here TDX continues the drift pattern that has been learned within the training window, but the drift pattern of the data changed, and possibly even reversed.\\ 
As can be taken from Fig. \ref{fig:wilcox} the MAE of the proposed method is only lower than that of the static model for the first nine forecasting timepoints, and only significant on 6 out of 9 compared to static. Furthermore TDX shows lower error than EDD for all forecasted time points with the differences being significant on 17 of the 25 time points.

\begin{figure}
\centering
\begin{minipage}{.48\textwidth}
  \centering
  \includegraphics[width=\linewidth]{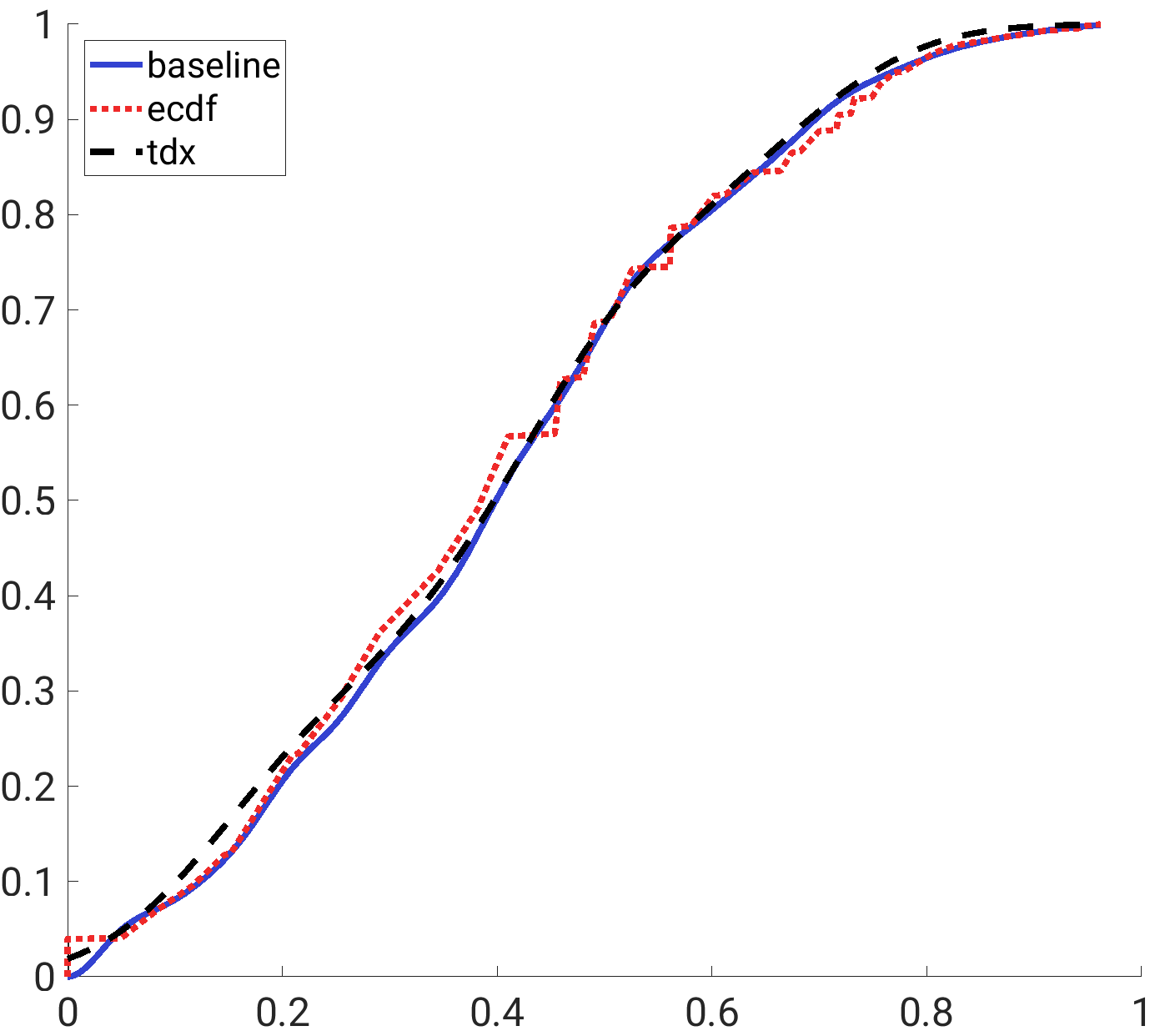} 
  \caption{CDF modelled by baseline (solid line), TDX (dashed line), as well as empirical CDF (dotted line). }\label{fig:int_rate_cdf_101}
\end{minipage}
\hfill
\begin{minipage}{.48\textwidth}
  \centering
     \includegraphics[width=\linewidth]{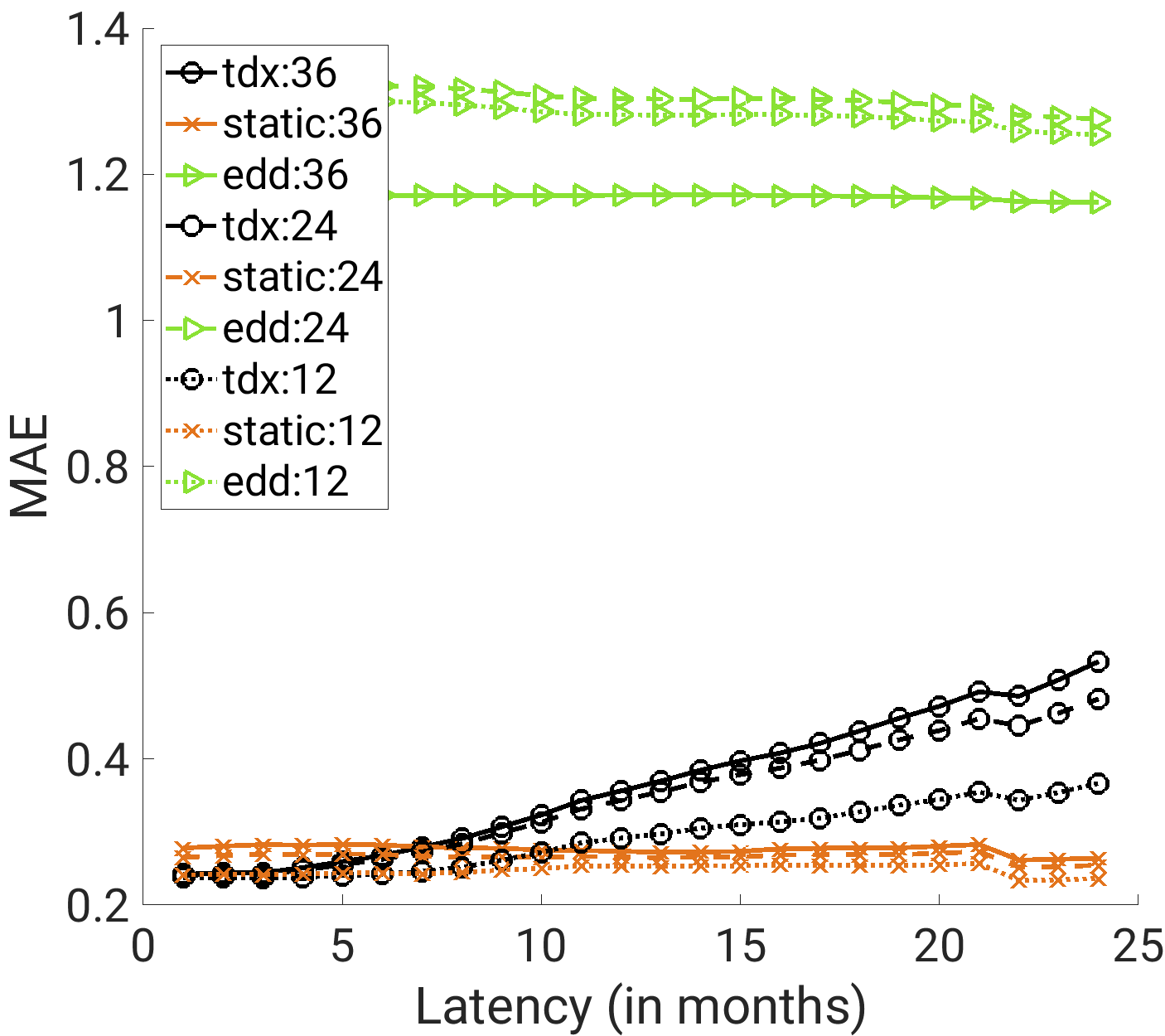} 
   \caption{Error development on the open\_acc data set}\label{fig:error_open_acc}

\end{minipage}%
\end{figure}

Fig. \ref{fig:int_rate_density_101} shows the baseline and predicted density on the int\_rate data with 5 months of latency. It is noticeable that the baseline density shows a lot of peaks and valleys compared to the revol\_util data set, which none of the methods seem to capture entirely. Inspection of the corresponding empirical cdf in Fig. \ref{fig:int_rate_cdf_101} shows that while there are deviations of the baseline from the ecdf, the baseline is reasonably close to the empirical distribution.

As shown in Fig. \ref{fig:wilcox} the results for the other three data sets from the lending club series, namely loan\_amnt, open\_acc and dti, show worse performance than the two previously presented. While TDX shows lower and significantly different error compared to EDD on all three and all forecasted time points, the performance of TDX compared to static is worse. For loan\_amnt and dti TDX shows consistently higher error than the static model. On open\_acc TDXs error is lower for the first 7 time points but insignificantly so. The density plots of these data sets included in Fig. \ref{fig:datadrift} show that the density at the later time points tends to only change slightly, as the darker lines are visibly close to each other. Similar to the results observed Fig. \ref{fig:error_staticskewnormals}, the static model proves as a tough competitor for TDX on these data sets. The MAE curves for these data sets are shown in Fig. \ref{fig:error_open_acc} as well as \ref{fig:error_loan_amnt} and \ref{fig:error_dti} (see appendix).

To summarise the results on the lending club data it can be said that TDX provides good forecasts on the two data sets that show a continuation of the drift pattern beyond the training window. If the drift does not continue or is not present at all, the static version of the model provides better estimates than the extrapolating one, which is to be expected.

\subsubsection{Pollution Data}\label{sec:results_pollution}
\begin{figure}
    \centering
    \begin{subfigure}[b]{0.48\textwidth}
        \includegraphics[width=\textwidth]{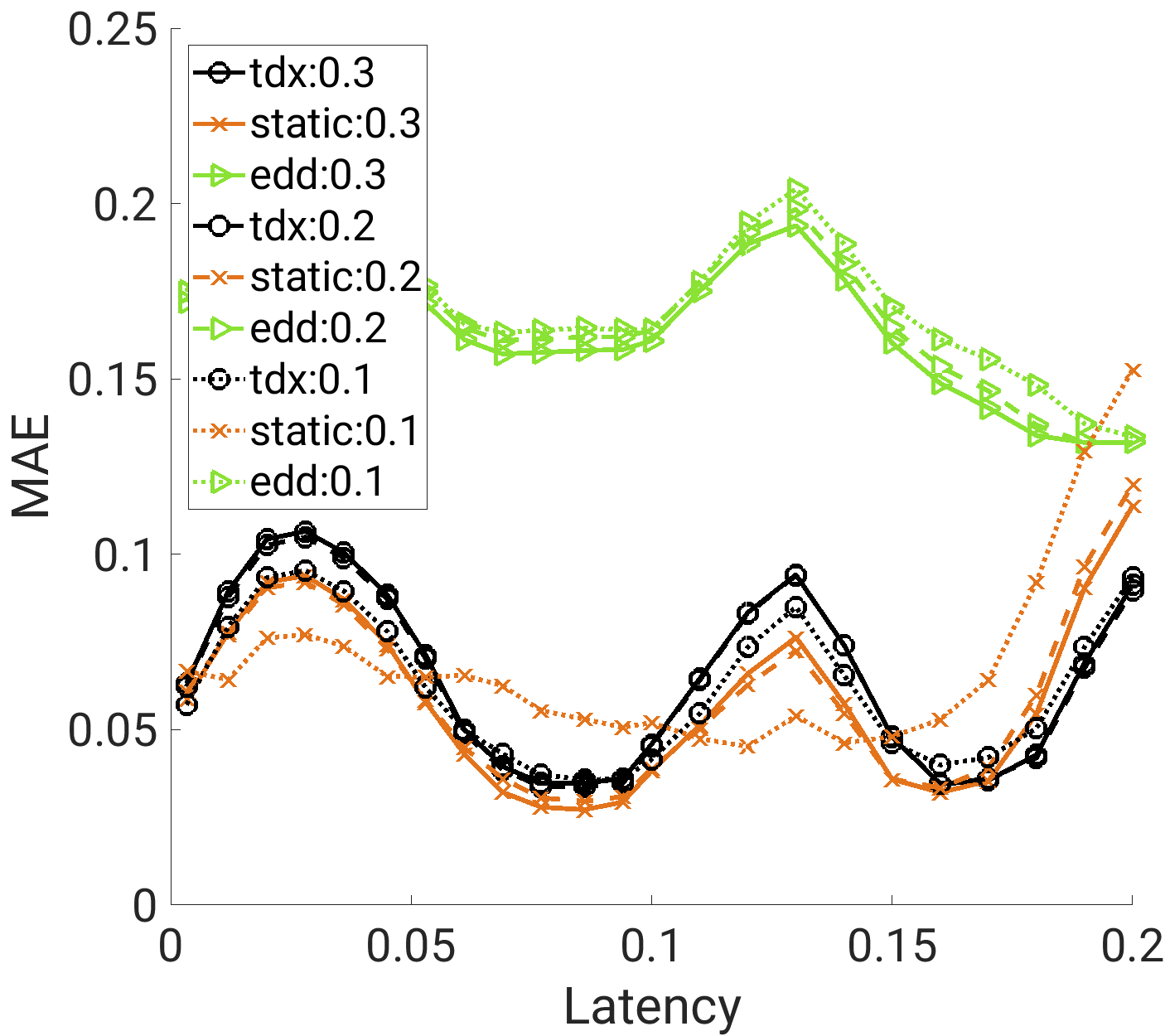} 
  \caption{}
  \label{fig:error_CO}

    \end{subfigure}
    ~ %
    \begin{subfigure}[b]{0.48\textwidth}
		\includegraphics[width=\textwidth]{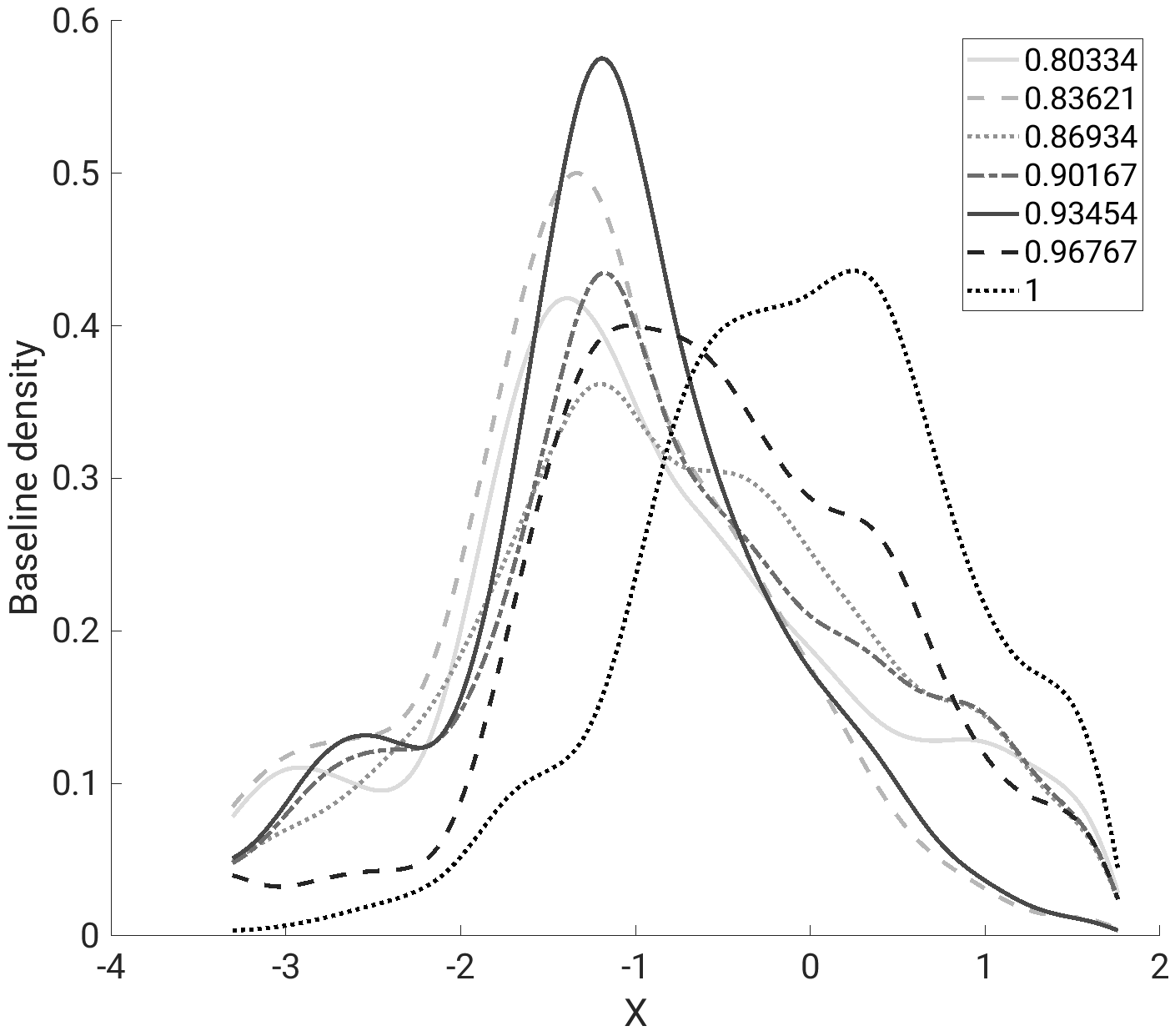} 
        \caption{}
        \label{fig:CO_density}
    \end{subfigure}
    \caption{(a): MAE of each method (indicated by colour) for each training window length (indicated by linestyle) for different latency values on the CO data set.
     (b): Baseline density of the CO data set within the test timeframe. Each line represents different time points as indicated in the legend.}
    \label{fig:PM10}
\end{figure}

The second series of real-world data sets is the Skopje pollution data. These data sets are an interesting addition to the previously presented ones as they not only show non-monotonous drift patterns, but these are to some extent repeating. This proves particularly challenging for all models in the experiments, since none of them is equipped to adjust to such repeating patterns.
An example of this can be seen in Fig. \ref{fig:error_CO} that shows the error on the CO data, while Fig. \ref{fig:CO_density} shows the development of the baseline density within the test timeframe. From the latter figure one can observe how the density around $x=-1.5$ increases, then decreases below the initial level, then increases again and again afterwards, before finally decreasing as the majority of the density shifts to the right. Matching this, an increase in error can be observed for all models around a latency of 0.03 and 0.13, right when the two major increases in density are observed. Due to the back-and-forth of this development, the TDX models struggle to anticipate this change, leading to the static model performing better overall by not trying to anticipate the movement of the drift. Nonetheless the TDX models manage to only score a slightly higher error for most of the forecasting time window and even managing to perform better at the end.  The test results in Fig. \ref{fig:wilcox} confirm this proximity, showing that the differences between TDX and static are only significant for only 14 of 25 time points.\\
While similar repeating patterns can be found on the other pollution data sets, these are not handled as well by TDX as on CO, resulting in the static method showing significantly lower error on large stretches of the forecasting time window on these data sets. Exceptions to this are observable on PM10 and PM25 between the time points 6 and 12, but it is nonetheless observable that the proposed method is not suited to handle this kind of drift patterns. The error curves for the O3, NO2, PM10 and PM25 data sets can be found in Fig. \ref{fig:error_pollution} in the appendix.

\subsection{Parameter Sensitivity Analysis}\label{sec:SA}
\begin{figure}
    \begin{subfigure}[b]{0.45\textwidth}
\includegraphics[width=\linewidth]{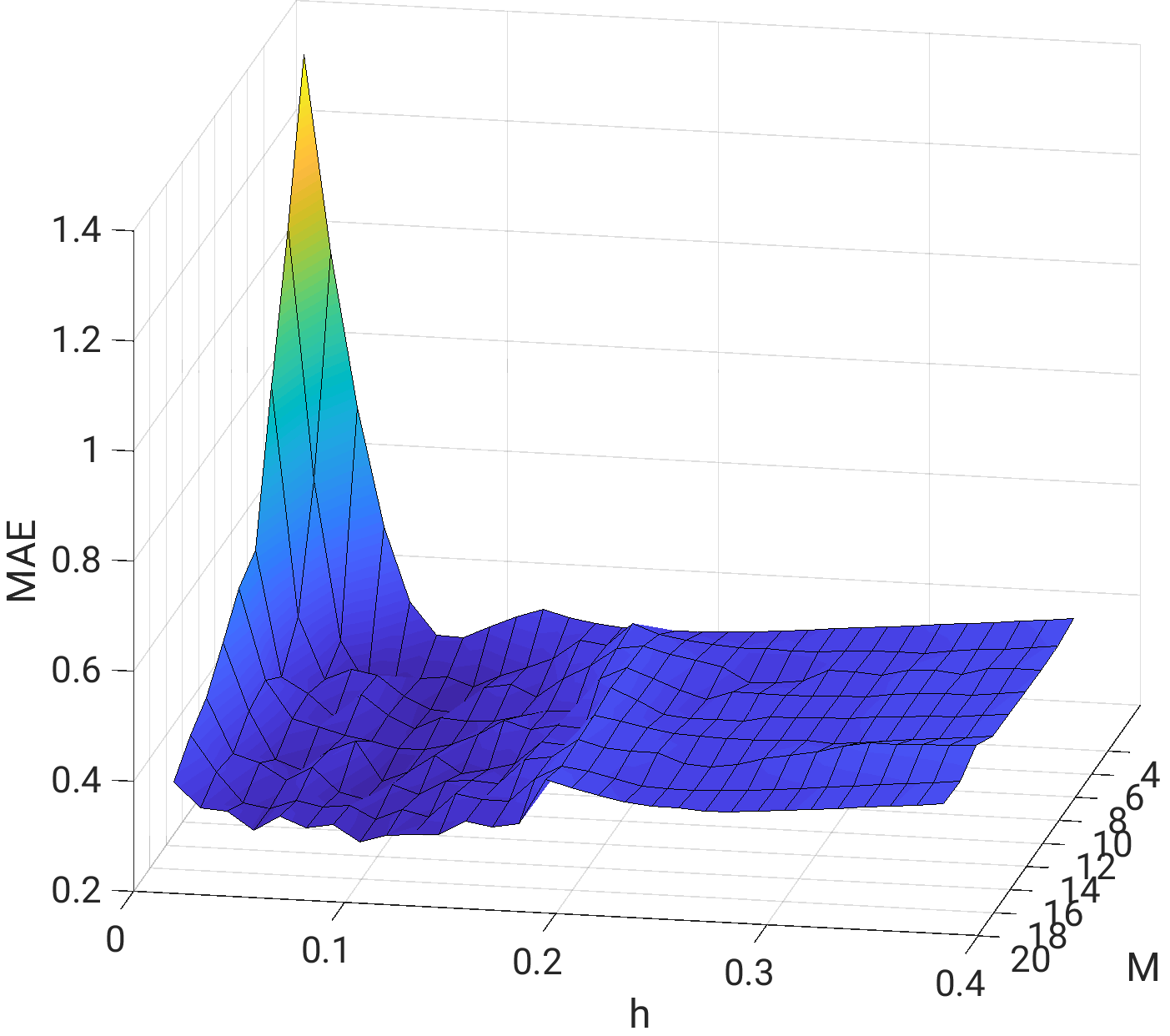} 
        \caption{int\_rate}
\label{fig:sa1_hm}
\end{subfigure}
    ~
    \begin{subfigure}[b]{0.45\textwidth}
\includegraphics[width=\linewidth]{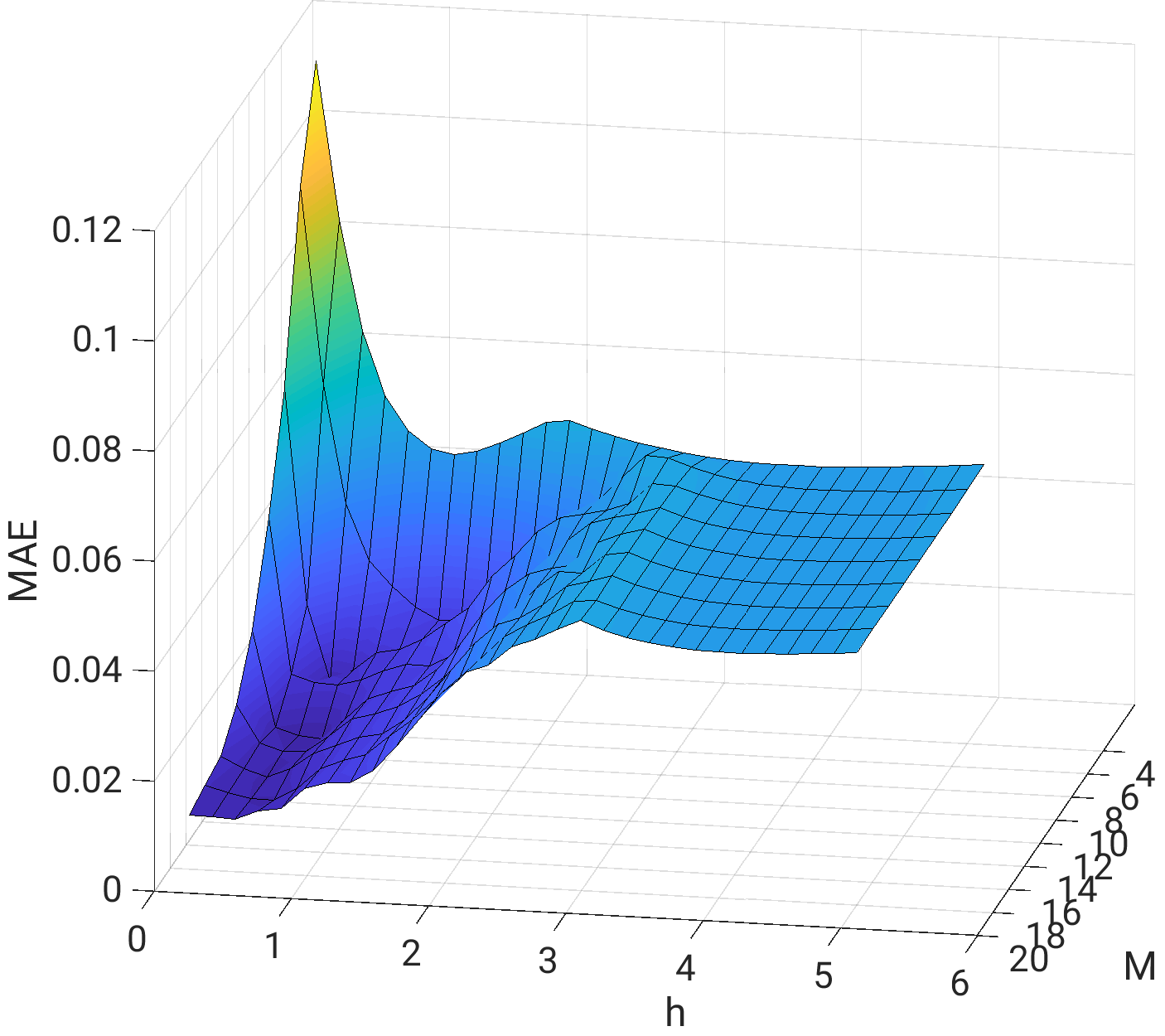} 
      \caption{meandrift}
      \label{fig:sa2_hm}
    \end{subfigure}
    \caption{Surface plots showing the MAE for different combinations of $M$ and $h$ values.}
\end{figure}

\begin{figure}
    \begin{subfigure}[b]{0.45\textwidth}
\includegraphics[width=\linewidth]{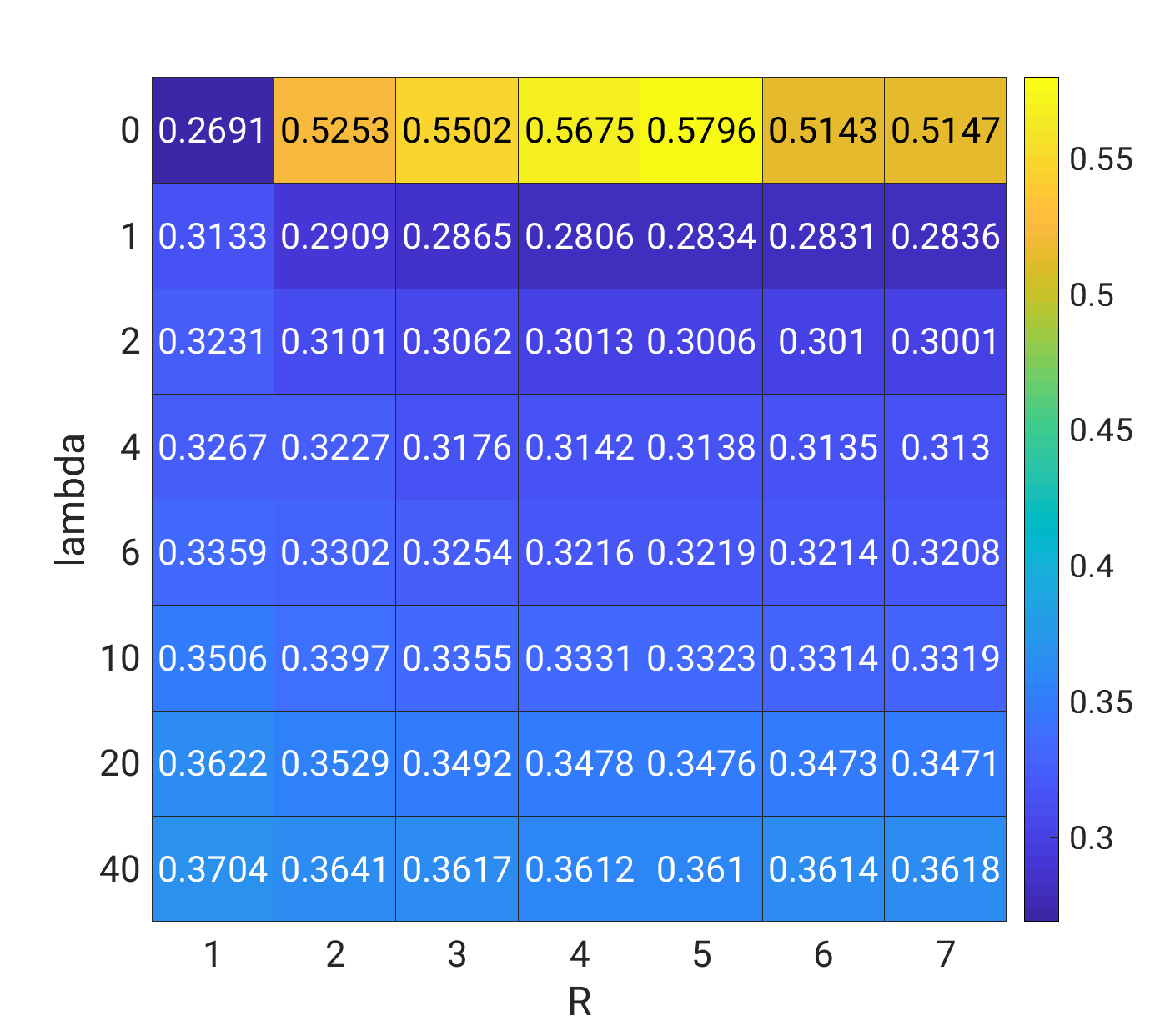} 
        \caption{int\_rate}
\label{fig:sa1_rl}
\end{subfigure}
    ~
    \begin{subfigure}[b]{0.45\textwidth}
\includegraphics[width=\linewidth]{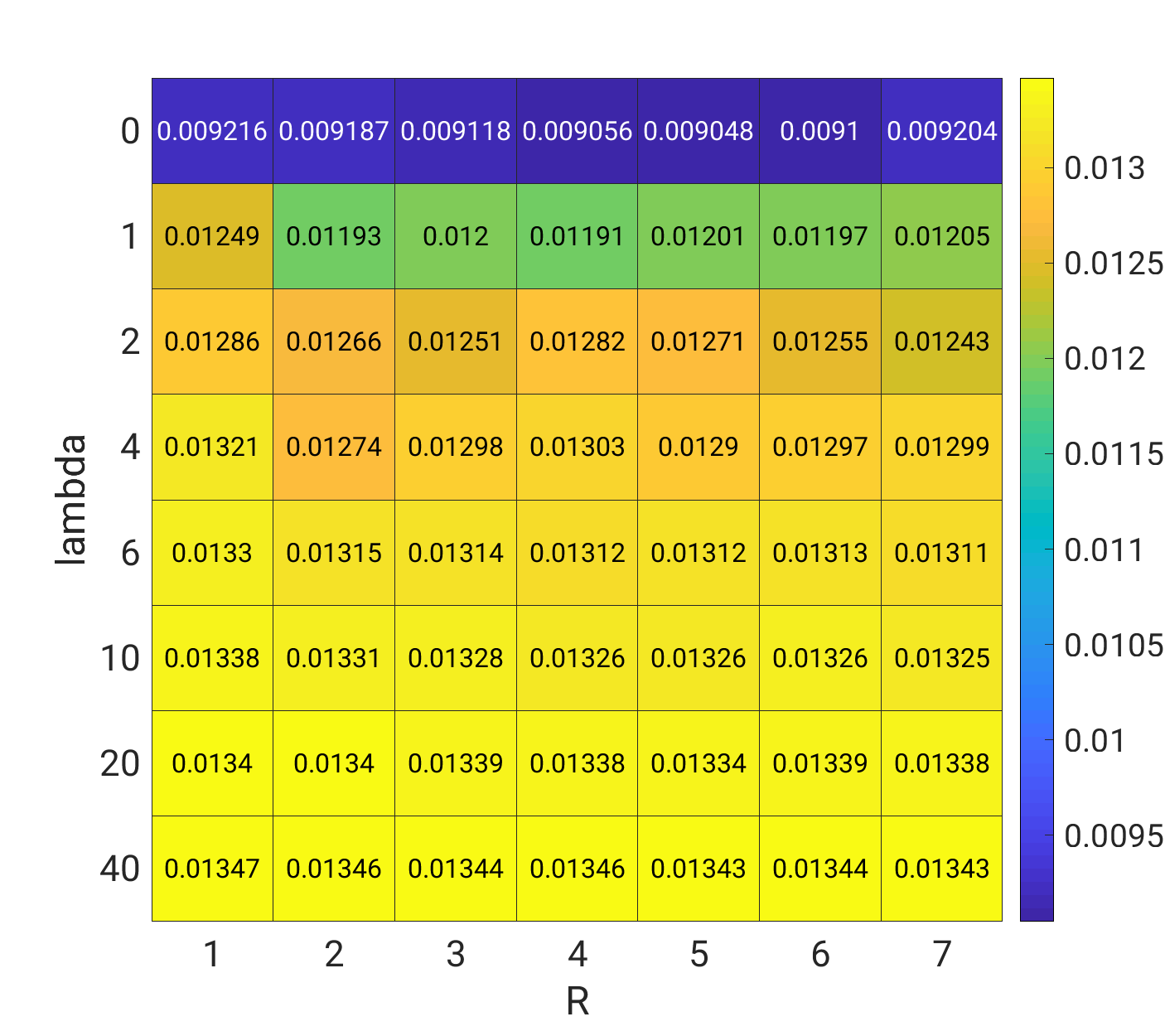} 
      \caption{meandrift}
      \label{fig:sa2_rl}
    \end{subfigure}
    \caption{Heatmaps showing the MAE for different combinations of $R$ and $\lambda$ values.}
\end{figure}

Since the proposed method has 4 hyperparameters we will discuss how sensitive the method is to changes in these parameters. To this end a series of experiments was conducted in the time interval of $t\in[0.3, 0.45]$ on the int\_rate and meandrift data sets. In the first phase of these experiments the number of basis functions $M$ and the bandwidth $h$ was investigated by spanning a parameter grid of 30 bandwidth values and 9 different numbers of bases, while keeping the other two parameters fixed at $R=2$ and $\lambda=1$. To compare the resulting models, the MAE of the model's prediction and the true/baseline density for latency of 0.05 was used.
In Fig. \ref{fig:sa1_hm} the MAE for different $h$ and $M$ values on the int\_rate data set is shown. While the extreme values for small $M$ and $h$ values make the surface difficult to read a region with lower error is visible for $h\leq0.2$.\\
The surface in Fig \ref{fig:sa2_hm} shows a clearer image for the results on the meandrift data set. There is a region with lower  error for $M$-values larger than 6 for $h$-values under 1, while very small $h$-values combined with small $m$-values result in higher error, reaching a maximum at the far corner of the surface  at $m=4$ and $h=0.25$. In this case the combination of few bases and small bandwidth strongly impacts the model performance. For $M$-values greater than 6 the error increases quite similarly with increasing $h$, eventually reaching a plateau at $h=3$ where the bandwidth is so large that the density estimate becomes so smooth that the influence of $M$ is no longer visible.
As it is also the case for the static density estimation, it can be seen that the choice of the $M$ and $h$ parameters is strongly dependent on the data so that it is best tuned on available data.\\
Proceeding with the best performing $M$ and $h$ values, the second phase of these experiments considers the influence of the remaining two hyperparameters, which are the order of the polynomial $R$ and the regularisation strength $\lambda$. The heatmap in Fig. \ref{fig:sa1_rl} shows the results for the int\_rate data set where the minimum lies at $R=6$ and $\lambda=1$. The results for $R>1$ and $\lambda=1$ form the region with the lowest errors, ranging from 0.26 to 0.28, while $R>1$ and $\lambda=0$ shows the highest error. The exact opposite is the case on the meandrift data set, whose results are shown in Fig. \ref{fig:sa2_rl}. Here, the region with lowest error is observed for $\lambda=0$, with the error increasing with increasing regularisation strength. In the case of meandrift, the drift pattern is very clear and shows little noise, leading to the unregularised models performing better. On real-world data like int\_rate the models benefit from a certain amount of regularisation, so $\lambda$-values around 1 appear to be a good choice as a default.


During all experiments above the training window was sufficiently large, so the number of used training samples was no concern. Fig. \ref{fig:sa3_ns} shows how the error behaves on meandrift for different numbers of samples used within the training window, employing the best parameters from Fig. \ref{fig:sa2_hm} and \ref{fig:sa2_rl} . In these experiments only every n-th instance was used, with $n \in \{124,62,31,$ $16,8,4,2,1\}$. This results in sample sizes ranging from 31 to 3744, the latter being all instances with $t\in[0.3, 0.45]$. The results show that while using the entirety of the samples resulted in the smallest error, using only around 500 instances achieved a comparable result. 

\subsection{Execution times}
\begin{figure}%
\centering
\begin{minipage}{.48\textwidth}
  \centering
 \includegraphics[width=\linewidth]{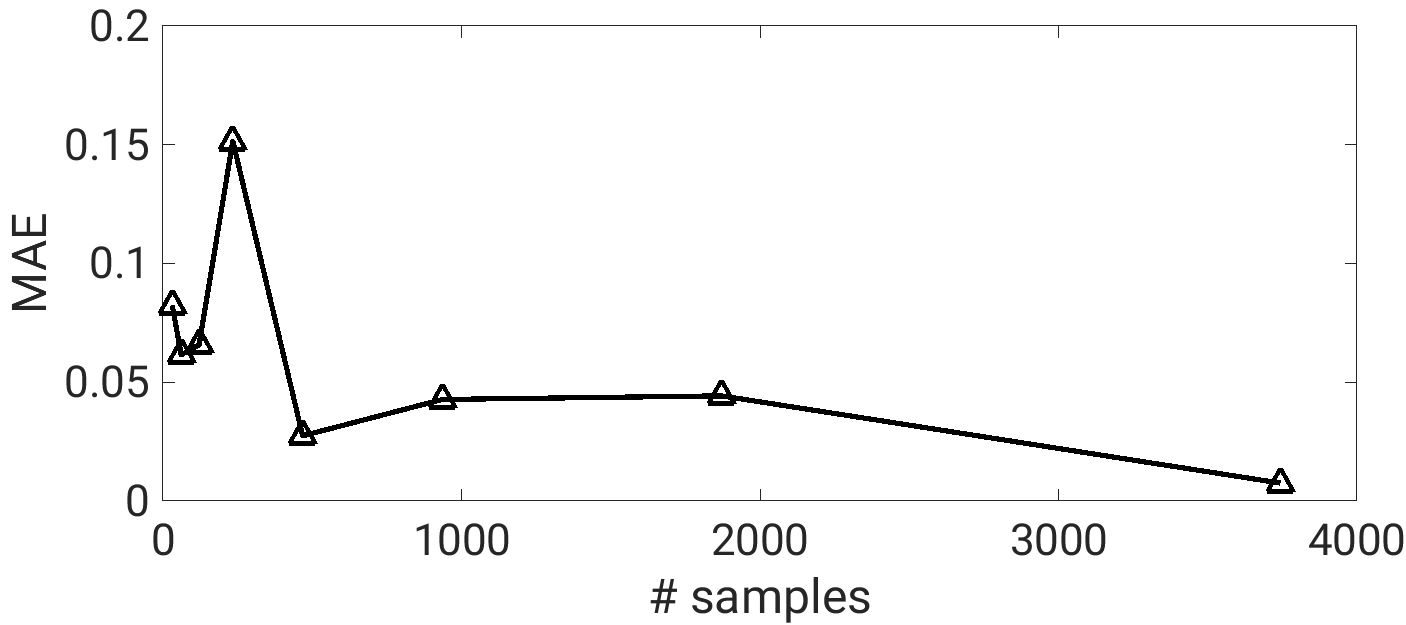} 
  \caption{MAE with varying number of samples taken from the training window}
  \label{fig:sa3_ns}
  \end{minipage}%
  \hfill
\begin{minipage}{.48\textwidth}
 \includegraphics[width=\linewidth]{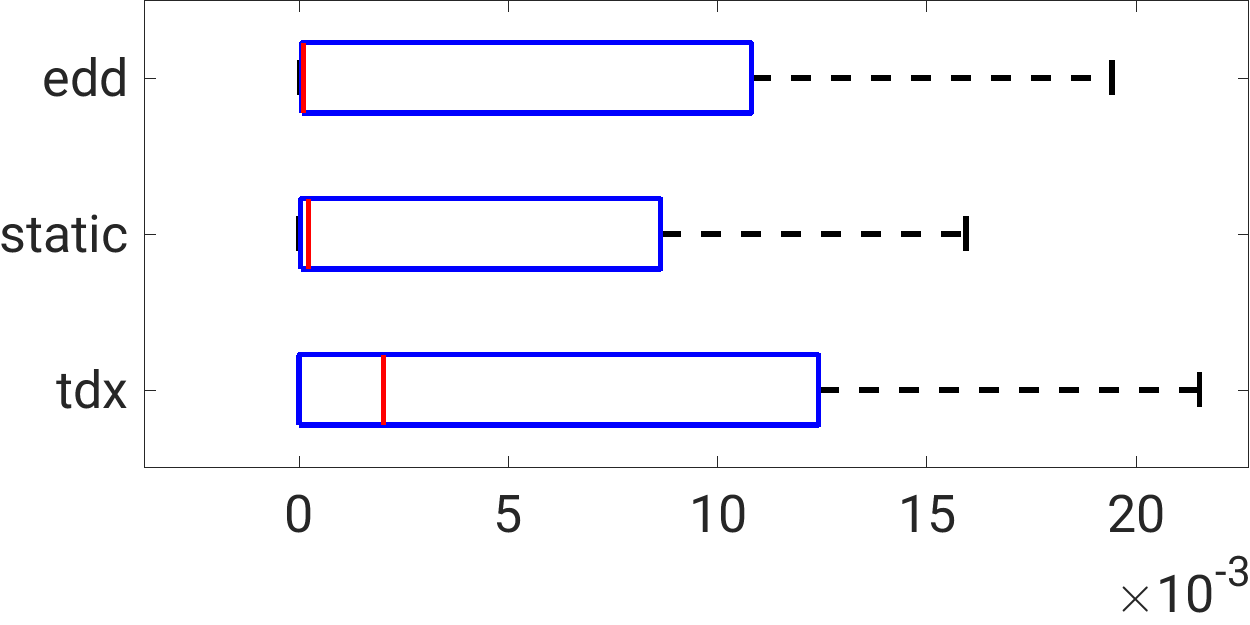} 
  \caption{Runtime in seconds required to fit the model divided by the number of training samples.}
  \label{fig:fittime}
\end{minipage}
\end{figure}

The execution time for fitting the model was measured programmatically during the experiments presented in Sec. \ref{sec:results_arti}-\ref{sec:results_pollution}. The boxplot in Fig. \ref{fig:fittime} shows the execution time in seconds the methods took to fit the model per observation in the training window. %
From the figure it is noticeable that TDX has the highest median fitting time, with the static model showing the lowest value for the 75th percentile (right end of box). Regarding the fitting time of TDX it has to be mentioned that during the experiments the optimisation problem is solved 4 times as part of a 'multistart' search as mentioned in Sec. \ref{sec:optprob}.  This configuration has been used as a reasonable default value and possibly has room for improvement.\\

\section{Conclusion} \label{sec:conc}
In this article we presented a novel approach called temporal density extrapolation (TDX) with the goal of predicting the probability density of a univariate feature in a data stream. TDX models the density as an expansion of Gaussian density basis functions, whose weights are modelled as functions of time to account for drift in the data. Fitting these time-dependent weighting functions is approached by modelling the weights of the basis expansion as compositional data. For this purpose the isometric log-ratio transformation is used to ensure the compliance with the properties of the basis expansion weights. This approach allows to extrapolate the density model to time points outside the available training window while accounting for the changes over time that are often encountered in data streams.\\
The evaluation shows that the TDX manages to capture monotonous drift patterns, like changes in the component means or weights of a mixture distribution, better than a competing method (EDD) or a static version of the basis expansion model. Furthermore the model also performs well on those two lending club data sets that exhibit very noticeable drift (revol\_util and int\_rate). All these data sets have in common that the drift in the data is both continuous and unidirectional. Data sets that show little to no drift however, like the static artificial data set or the remaining lending club data sets, prove challenging for TDX. In these cases the static version of the model that  fits the weights of the basis expansion in a non-time-adaptive fashion performs better, as the data generating process aligns with its stationarity assumption. \\
Furthermore the results on the pollution data sets show that the method is currently not equipped to handle drift patterns that show seasonality. As the model is designed to handle monotonous drift, this is not surprising but shows another avenue for future work.\\
The analysis of the model's sensitivity with regard to its hyperparameters has shown that both the number of basis functions $M$ and their bandwidth $h$ has to be tuned on the data, as is necessary for the static density estimation model.
For both extrapolation-specific parameters there exist reasonable defaults: for $R$, the order of the polynomial used to model the basis weights, and the regularisation strength $\lambda$. This reduces the effort of configuring TDX. \\
In summary, TDX handles its intended application, i.e., data with monotonous drift, better than the only other comparable density forecasting approach to date (EDD) and provides reliable density forecasts on these data sets. The next step in the models development will be the use of TDXs density forecasts for probabilistic classification in data streams. Furthermore we aim to extend the methods capabilities by improving the performance on data sets with little to no drift. 
\smallskip \\ \textbf{Acknowledgements} \newline
We thank the Austrian National Bank for supporting our research project number 17028 as part of the OeNB Anniversary Fund. We thank  Christoph Lampert for providing his implementation of the EDD approach and Utrecht University for providing the Gemini cluster for computations.

\pagebreak[3]
\bibliography{dami_temporal_density_extrapolation}
\bibliographystyle{spbasic} 

\pagebreak[3]
\clearpage{}%
\section*{Appendix}
This section contains additional material that was excluded from the main part of the work out of consideration of the available space as well as the information flow. \\
In the following the derivation of the objective gradient can be found. This is relevant because it is used to solve the optimization problem entailed by proposed method. \\
After that a table detailing the hyperparameters of both TDX and EDD that resulted from the model selection procedure can be found. Note that the static version of TDX is not listed separately, as it uses the same $M$ and $h$ parameters as the adaptive TDX version.

\bigskip
\textbf{Model Selection Results}\\
\bigskip

\begin{table}[!ht]
\begin{center}
\begin{tabular}{r|cccc|cc}
 &\multicolumn{4}{|c|}{TDX}          & EDD        \\ \hline
Data Set & M & h & R & $\lambda$ & $\sigma$ & $\lambda$\\ \hline
meandrift & 14  & 0.62  & 2  & 1  & 0.3 & 0.028     \\
weightdrift & 14  & 0.42  & 2  &  1     &  0.13 & 0.004  \\                        
sigmachange & 10 & 0.46  &  3 &  2   &  0.3 & 0.004  \\                       
staticskewnormals &  14 & 0.53  & 1  &    4  & 0.15 & 0.004 \\ \hline                          
dti & 10  &  0.007 & 3  &  1      &     0.095 & 0     \\               
int\_rate & 10  & 0.091  & 3  &   1    &  0.14 & 0     \\                     
loan\_amnt & 12  &  0.056 & 3  & 3     &   0.048 & 0   \\                     
open\_acc &  10 &  0.037 &  3 &  1    &    0.11  & 0     \\                  
revol\_util & 14  &  0.007 & 3  &  1    &  0.23 & 0  \\\hline                         
CO &  12 & 0.38  & 1  & 5           &      0.3   & 0.0014     \\               
NO2 &  12 & 0.043  & 3  &  1          &     0.0216 & 2.8e-5          \\              
O3 & 10  & 0.075  & 1  &   1         &     0.0629 & 1.4e-5           \\             
PM10 & 10  & 0.42  & 1  &  5          &      0.3  & 0.0012      \\               
PM25 &  10 & 0.044  & 1  &  5          &    0.15 & 4.6e-5     \\              
\end{tabular}
\end{center}
\caption{Hyperparameters of the models resulting from the model selection procedure}
\label{tab:parameters}
\end{table}

\clearpage
\bigskip
\textbf{Additional Figures}\\
\vspace{-1cm}
\begin{figure}[!htb]
    \centering
    \begin{subfigure}[b]{0.48\textwidth}
    	\includegraphics[width=\textwidth]{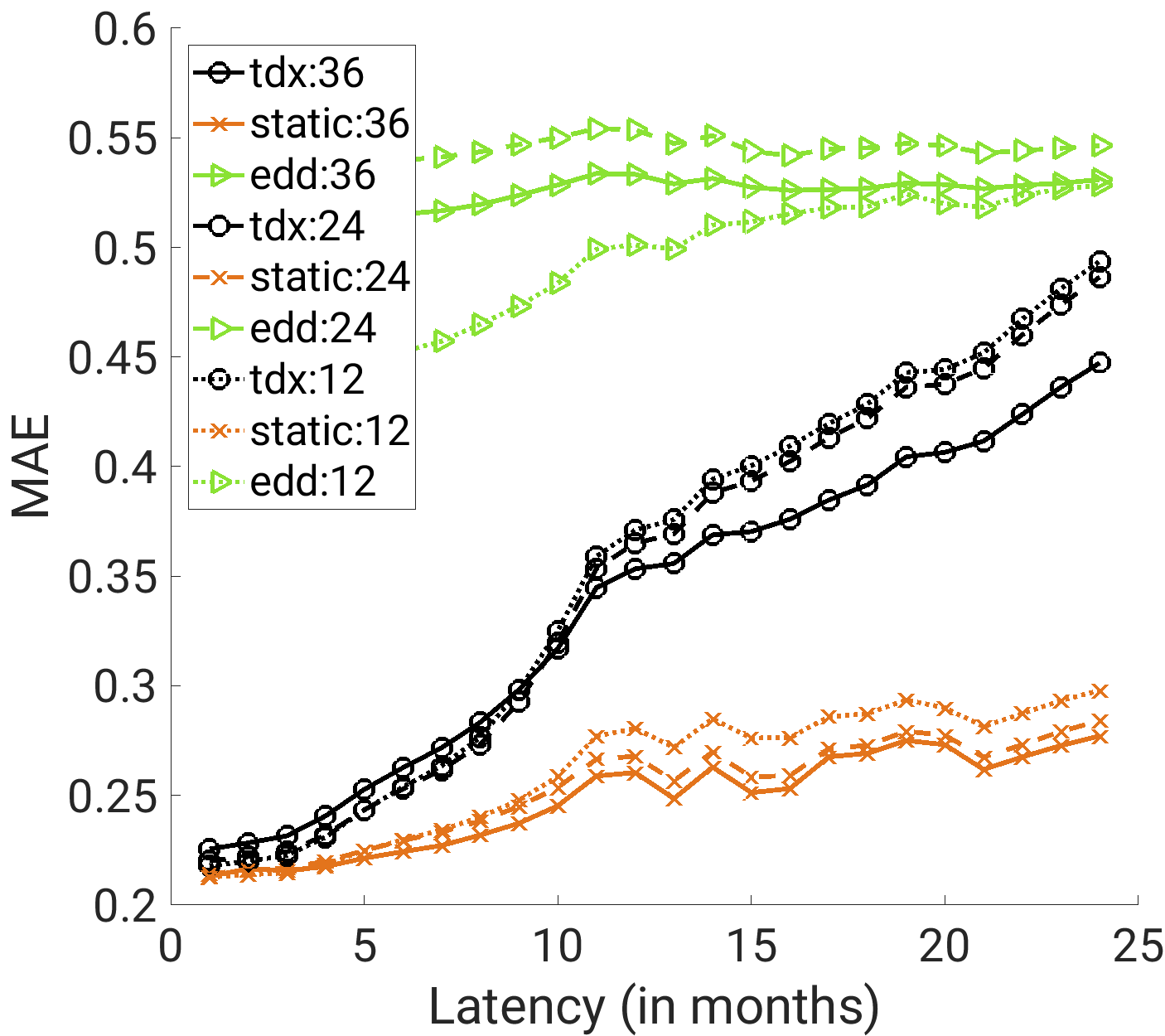} 
    	\caption{loan\_amnt}
    	\label{fig:error_loan_amnt}
    \end{subfigure}
    ~ %
    \begin{subfigure}[b]{0.48\textwidth}
        \includegraphics[width=\textwidth]{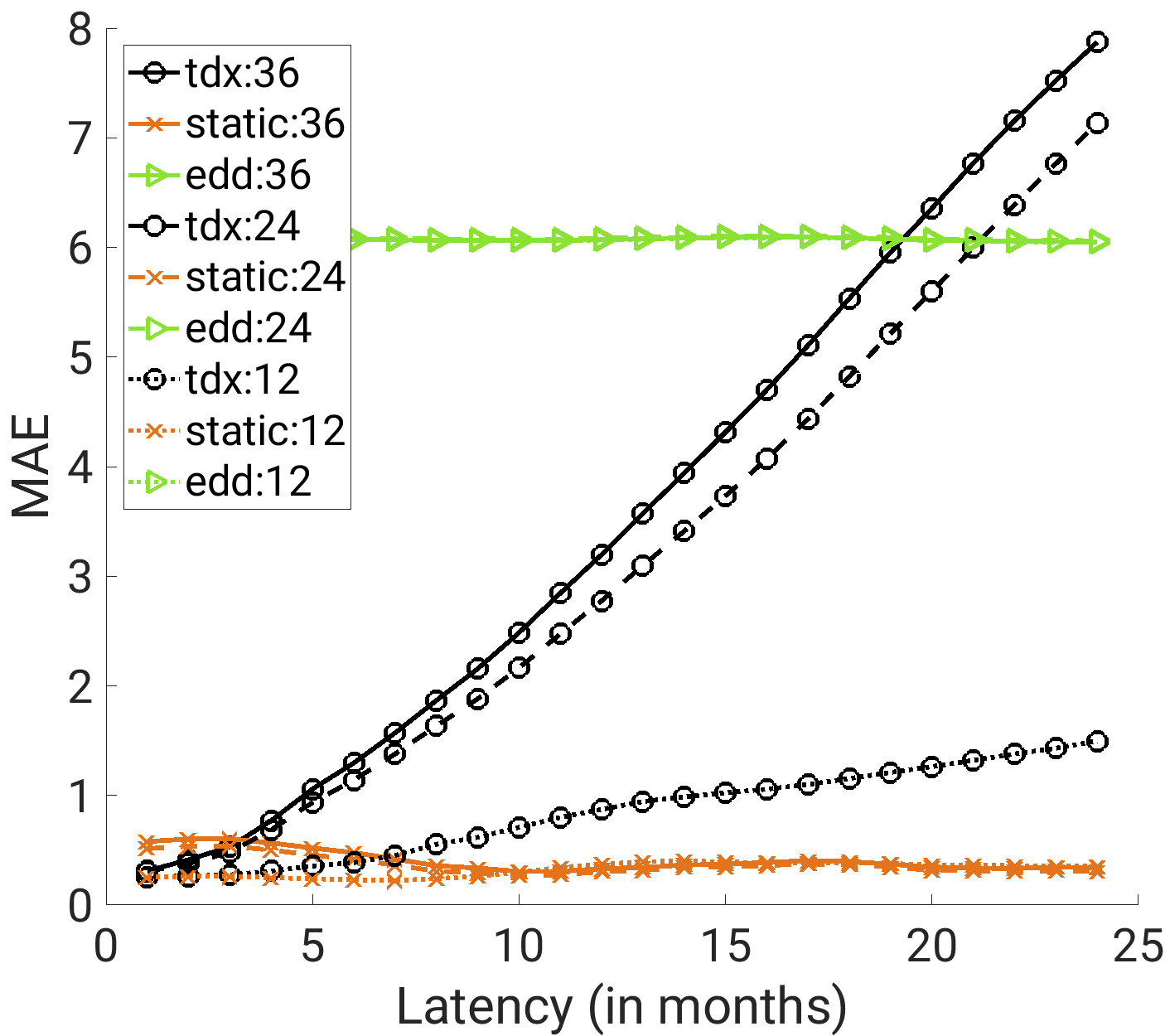} 
        \caption{dti}
        \label{fig:error_dti}
    \end{subfigure}
    
    \begin{subfigure}[b]{0.48\textwidth}
        \includegraphics[width=\textwidth]{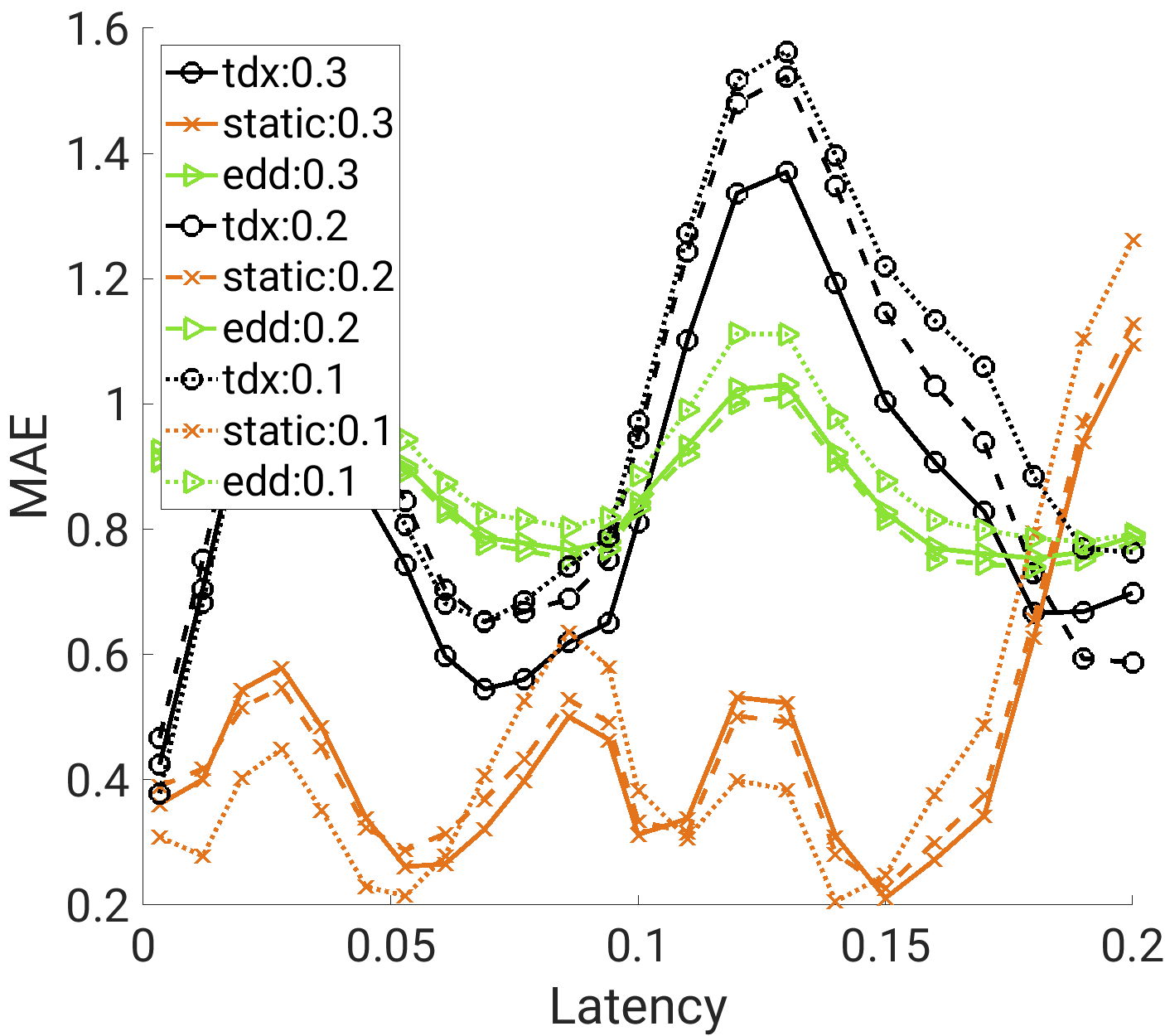} 
        \caption{O3}
        \label{fig:error_O3}
    \end{subfigure}
    ~ %
    \begin{subfigure}[b]{0.48\textwidth}
        \includegraphics[width=\textwidth]{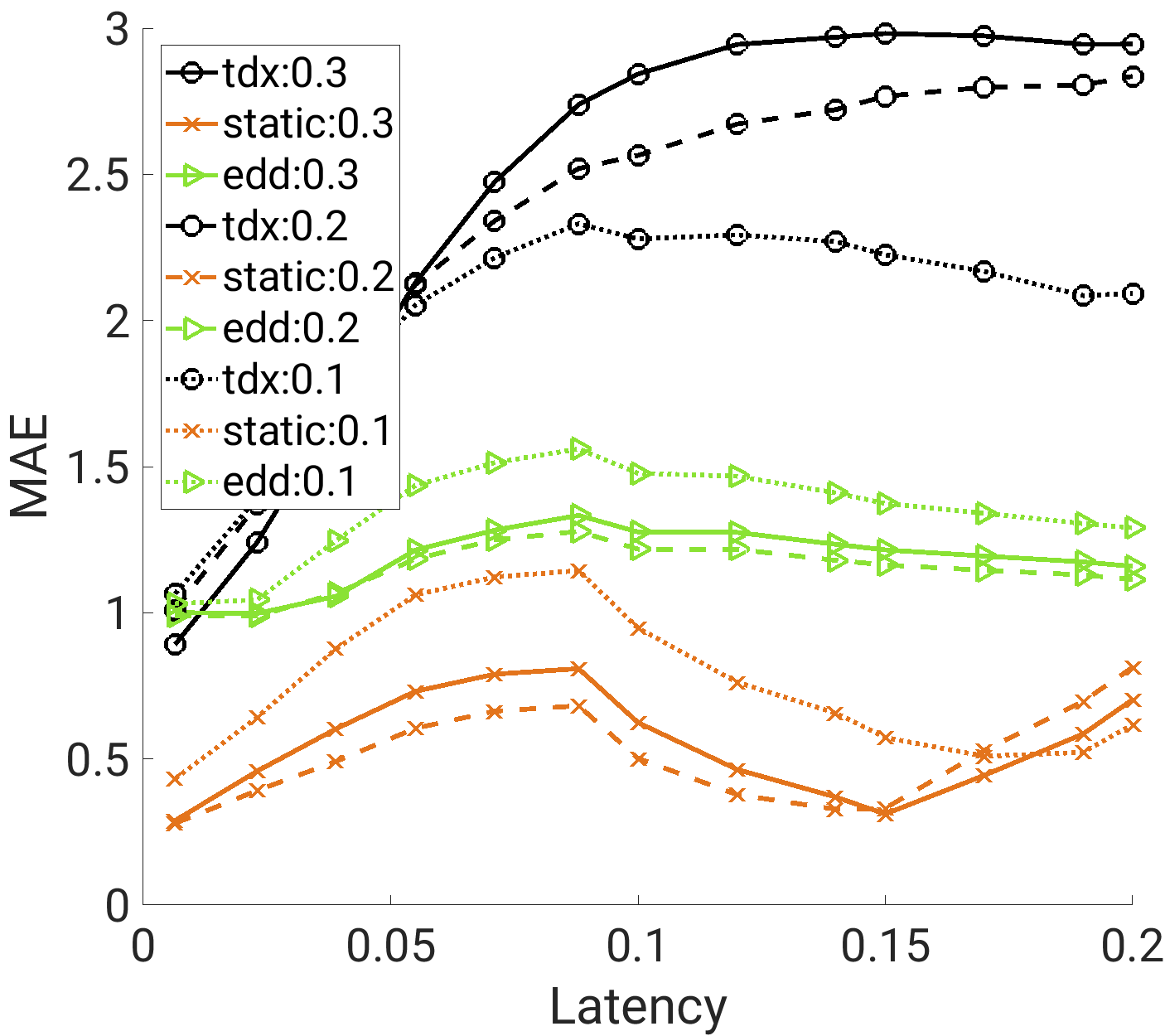} 
        \caption{NO2}
        \label{fig:error_NO2}
    \end{subfigure}
    
    \begin{subfigure}[b]{0.48\textwidth}
        \includegraphics[width=\textwidth]{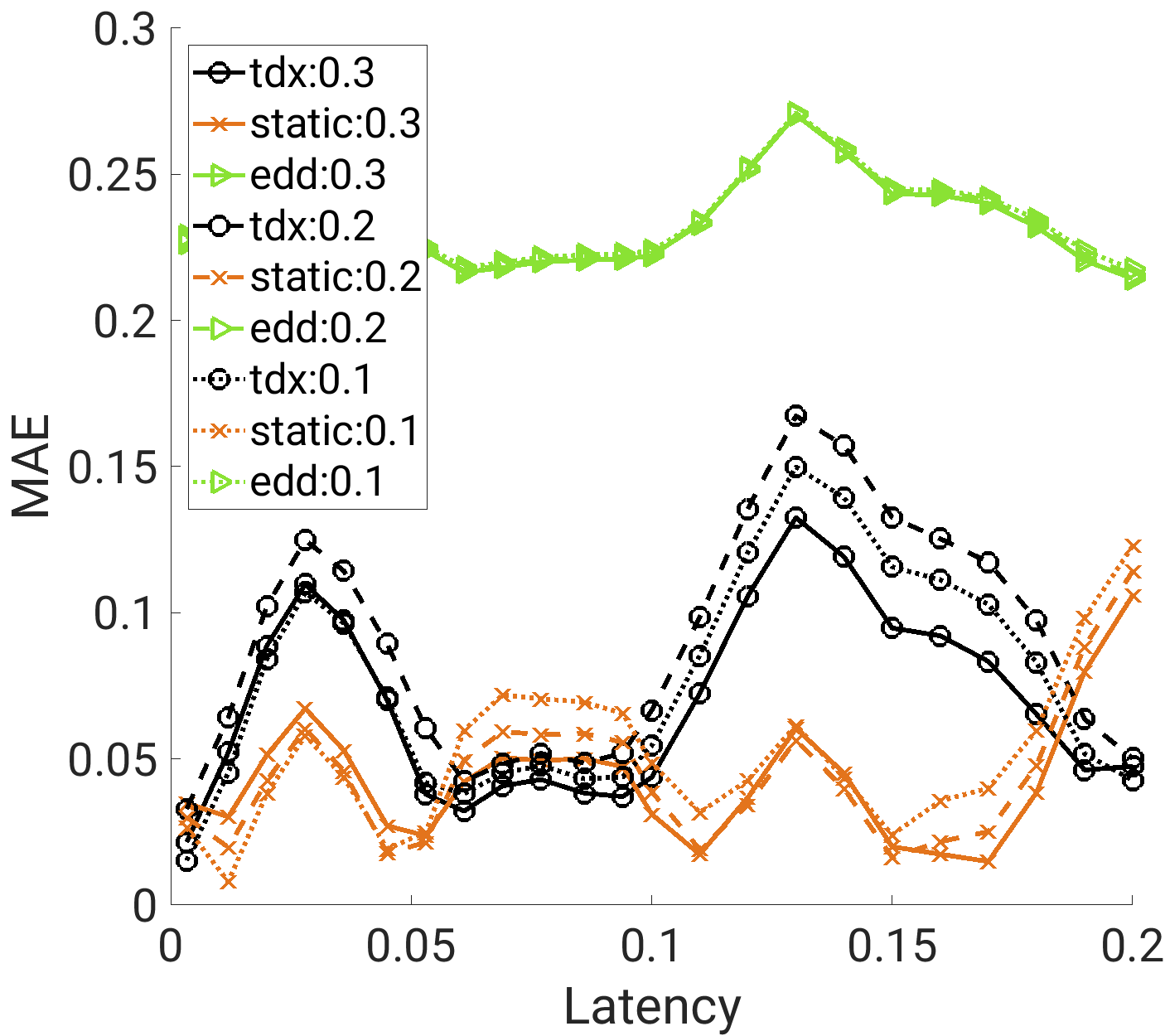} 
        \caption{PM10}
        \label{fig:error_PM10}
    \end{subfigure}
    ~
    \begin{subfigure}[b]{0.48\textwidth}
      \includegraphics[width=\textwidth]{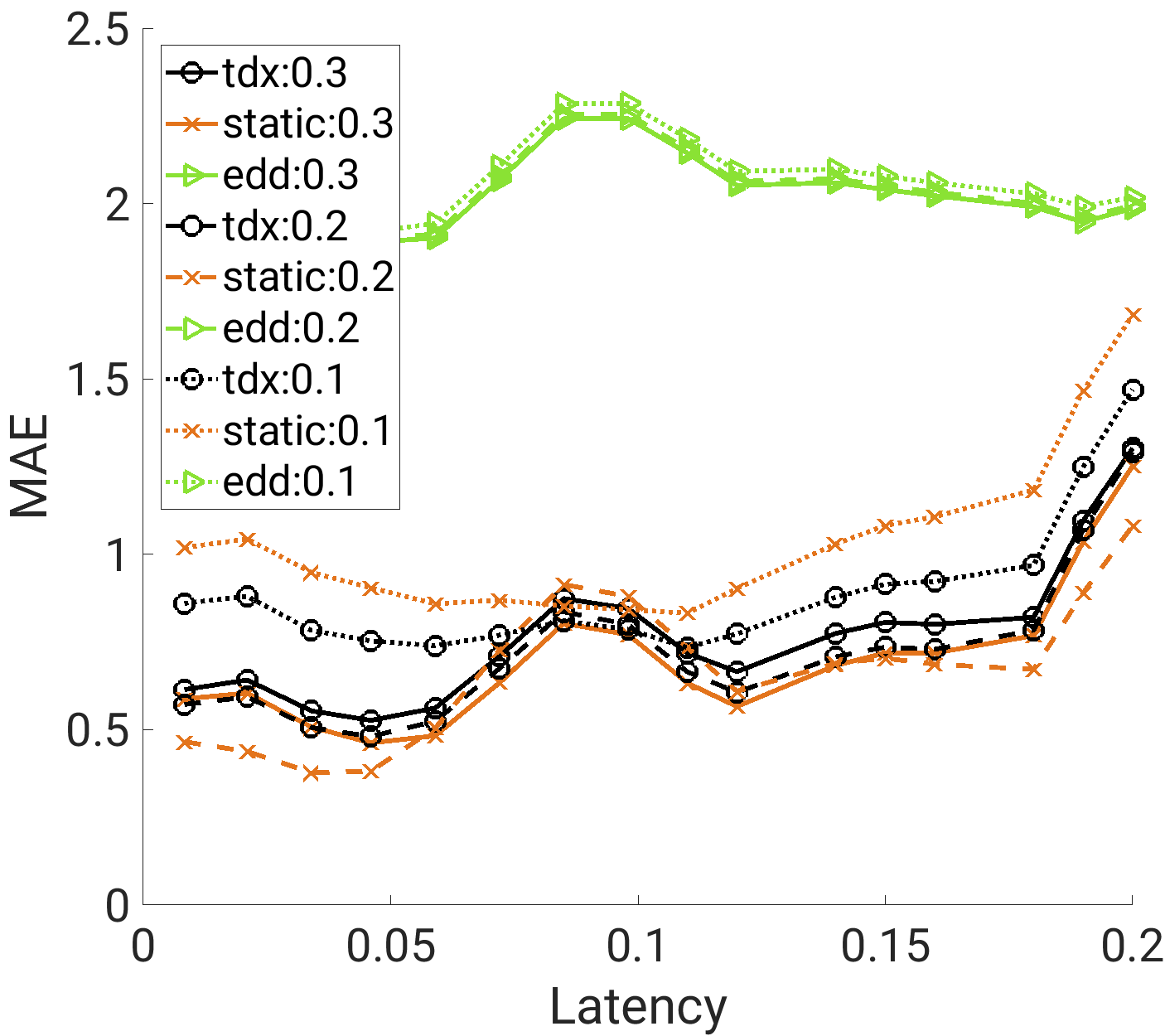} 
      \caption{PM25}
      \label{fig:error_PM25}
    \end{subfigure}
    \caption{MAE of each method (indicated by colour) for each training window length (indicated by linestyle) for different latency values on the remaining lending club and pollution data sets.}\label{fig:error_pollution}
\end{figure}

\clearpage

\textbf{Derivation of the Objective Gradient}\\
\bigskip

Consider the log-likelihood ($l$) equation of the temporal density extrapolation model, excluding instance weighting and regularization.

\begin{equation}
l=\sum_{i=1}^N\log(\bm{\phi}(x_i)^\intercal\exp(\bm{U}\bm{B}\bm{a}(\tau_i)))-\sum_{i=1}^N\log(\bm{1}^\intercal\exp(\bm{U}\bm{B}\bm{a}(\tau_i))\label{eq:loglikelihood_new_ap}
\end{equation}

Let us define a generalized version of the terms in Eq. \ref{eq:loglikelihood_new_ap} as $g(\bm{y})$ 

\begin{equation}\label{eq:gfun_ap}
	g(\bm{y}) =  \sum_{i=1}^n\log(\bm{y}^\intercal\exp(\bm{U}\bm{B}\bm{a}(\tau_i))).
\end{equation}

Then the log-likelihood in Eq. \ref{eq:loglikelihood_new_ap} and its gradient $\nabla l$ can be expressed as
\begin{align}
	l&=g(\bm{\phi}(x_i)) - g(\bm{1}),\\
	\nabla l &= \nabla g(\bm{\phi}(x_i)) - \nabla g(\bm{1}). \label{eq:grad_sub_ap}
\end{align}

As such we will start by deriving the gradient of $g(\bm{y})$. Recall the matrix $\bm{U}$ required for the ilr-transformation

\begin{align}
\bm{U}&=\widetilde{\bm{U}}\cdot \bm{D}_2\label{eq:u}\\
\widetilde{\bm{U}}&=\begin{pmatrix*}[r]
-1&-1&-1&\cdots&-1\\
1&-1&-1&\cdots&-1\\
0&2&-1&\cdots&-1\\
0&0&3&\cdots&-1\\
\vdots&\vdots&\vdots&\ddots&\vdots\\
0&0&0&\cdots&M-1\\
\end{pmatrix*}=(\widetilde{\bm{u}}_1,\ldots,\widetilde{\bm{u}}_{M-1})\label{eq:utilde}\\
    \bm{D}_2&=
\begin{pmatrix}
\frac{1}{\Vert\widetilde{\bm{u}}_1\Vert_2}&0&0&\cdots&0\\
0&\frac{1}{\Vert\widetilde{\bm{u}}_2\Vert_2}&0&\cdots&0\\
\vdots&\vdots&\ddots&\cdots&\vdots\\
0&0&0&\cdots&\frac{1}{\Vert\widetilde{\bm{u}}_{M-1}\Vert_2}
\end{pmatrix}\label{eq:d2_ap}
\end{align}

which is part of the term $\bm{U} \bm{B} \bm{a} (\tau_i)$ appearing in both parts of $l$. This term is then denoted as $\bm{\kappa}$ and its Jacobian matrix with respect to $B$ is denoted as $\bm{J}$.

\begin{eqnarray}
	\bm{\kappa} &=& \bm{U} \bm{B} \bm{a} (\tau_i)\\
	\frac{\partial \bm{\kappa}}{\partial \bm{B}} &=& \bm{J}
\end{eqnarray}

As $\bm{U} \bm{B} \bm{a} (\tau_i)$ appears in Eq. \ref{eq:gfun_ap} in exponentiated form, we define

\begin{eqnarray}
	\bm{E}&=&exp(\bm{\kappa})=
	\begin{pmatrix}
	exp(\kappa_1)\\
	\vdots\\
	exp(\kappa_N)\\
	\end{pmatrix}=
	\begin{pmatrix}
	\varepsilon_1\\
	\vdots\\
	\varepsilon_N\\
	\end{pmatrix}\\
\end{eqnarray}

and derive it in the following

\begin{eqnarray}
	\frac{\partial \varepsilon_i}{\partial \bm{B}_{ij}} &=& exp(\kappa_i)\frac{\partial \kappa_i}{\partial \bm{B}_{ij}}\\
	&=& exp(\kappa_i) \bm{U}_{ij} \bm{a}_k\\
	\bm{D} &=& diag(\bm{E})= \begin{pmatrix}
	\varepsilon_1 & 0 &\cdots&0\\
	0 & \varepsilon_2  &\cdots&0\\
	\vdots& \vdots &\ddots & \vdots\\
	0 & \cdots & \cdots & \varepsilon_N
	\end{pmatrix}\\
	\frac{\partial \bm{E}}{\partial \bm{B}} &=& \bm{D} \cdot \bm{J}
\end{eqnarray}

Then we reintroduce the placeholder variable $\bm{y}$ and define $\beta$ as the expression inside the logarithm in Eq. \ref{eq:gfun_ap}

\begin{eqnarray}
	\bm{y}^\intercal \bm{E} &=& \bm{y}^\intercal\cdot exp(\bm{\kappa}) \\
	&=& \bm{y}^\intercal \cdot exp(\bm{U} \bm{B} \bm{a}) = \beta  \\
\end{eqnarray}
and take the derivative of this expression.
\begin{eqnarray}
	\frac{\partial \beta}{\partial \bm{B}_{ij}} &=& \bm{y}^\intercal \cdot \frac{\partial \bm{E}}{\partial \bm{B}_{ij}} \\
	\frac{\partial \beta}{\partial \bm{B}} &=&\bm{y}^\intercal \bm{D} \bm{J}
\end{eqnarray}

Returning to the formulation in Eq. \ref{eq:gfun_ap}, we arrive at its gradient $\nabla g$ by by substituting with the derivatives of its parts shown above.

\begin{eqnarray}
	g &=& log(\beta) = log(\bm{y}^\intercal \bm{E})\\
	\frac{\partial g}{\partial \bm{B}} &=& \frac{1}{\beta} \bm{y} \bm{D} \bm{J}\\
	\nabla g &=&  \frac{1}{\beta} \bm{y} \bm{D} \bm{J}
\end{eqnarray}

Applying this to the original formulation in Eq. \ref{eq:grad_sub_ap} we get

\begin{eqnarray}
	\nabla l &=& \sum_{i=1}^{N}(\nabla g(\bm{\phi}(x_i)) - \nabla g(\bm{1}))\\
	\nabla l &=& \sum_{i=1}^{N}((\frac{1}{\beta} \bm{\phi}(x_i) \bm{D} \bm{J} ) - (\frac{1}{\beta} \bm{1} \bm{D} \bm{J} )). \label{eq:gradient_ap}
\end{eqnarray}

The inclusion of the temporal instance weighting via a weight vector $\bm{w}$ is then straight-forward.

\begin{eqnarray}
l&=&\sum_{i=1}^N w_i\log(\bm{\phi}(x_i)^\intercal\exp(\bm{U}\bm{B}\bm{a}(\tau_i)))-\sum_{i=1}^N w_i \log(\bm{1}^\intercal\exp(\bm{U}\bm{B}\bm{a}(\tau_i))\\\\
\nabla l &=& \sum_{i=1}^{N} w_i  (\frac{1}{\beta} \bm{\phi}(x_i) \bm{D} \bm{J})  - \sum_{i=1}^{N} w_i  (\frac{1}{\beta} \bm{1} \bm{D} \bm{J} ))
\end{eqnarray}

This leaves only the regularization term $\zeta$ out, which was defined as

\begin{align}
\zeta =  \lambda \ tr(\bm{C}^\intercal \bm{B}^\intercal \bm{B} \bm{C}) \label{eq:regularization_ap}\\
\bm{C} = \begin{pmatrix}
0&0&\cdots&0\\
1 & 0&\cdots &0\\
0&1&\cdots&0\\
\vdots&\vdots&\ddots&\vdots\\
0&0&\cdots&1
\end{pmatrix}
\end{align}

whose derivative then is $\lambda \ \bm{B} \bm{C} \bm{C}^\intercal$, arriving at the final form of the gradient

\begin{equation}
\nabla l =( \sum_{i=1}^{N} w_i  (\frac{1}{\beta} \bm{\phi}(x_i) \bm{D} \bm{J})  - \sum_{i=1}^{N} w_i  (\frac{1}{\beta} \bm{1} \bm{D} \bm{J} )) ) - \lambda \ \bm{B} \bm{C} \bm{C}^\intercal
\end{equation}

\clearpage{}%

\end{document}